\documentclass[manuscript,screen]{acmart}
\AtBeginDocument{%
  \providecommand\BibTeX{{%
    \normalfont B\kern-0.5em{\scshape i\kern-0.25em b}\kern-0.8em\TeX}}}

\setcopyright{acmcopyright}
\copyrightyear{2023}
\acmYear{2023}
\acmDOI{XXXXXXX.XXXXXXX}

\usepackage{balance}       
\usepackage{graphics}      
\usepackage{hyperref}
\usepackage{color}
\usepackage{booktabs}
\usepackage{textcomp}
\usepackage{subcaption}
\usepackage{enumerate}
\usepackage{xcolor}
\usepackage{lipsum}
\usepackage{makecell}
\usepackage{multicol}
\usepackage{multirow}
\usepackage{array}
\usepackage{verbatimbox}
\usepackage{enumitem}
\usepackage{amsmath}
\usepackage{stfloats}
\usepackage{graphicx}
\usepackage{amsthm}
\usepackage{listings}
\usepackage{caption} 
\usepackage[export]{adjustbox}
\usepackage{xspace}
\usepackage{epsfig}
\usepackage[linesnumbered]{algorithm2e}
\usepackage{algpseudocode}
\usepackage{tabularx}
\usepackage{arydshln}
\usepackage[bottom]{footmisc}
\usepackage{tcolorbox}
\usepackage{stackengine}
\usepackage{placeins}
\usepackage[normalem]{ulem}

\newcommand*{\eg}{\textit{e.g.},\xspace}
\newcommand*{\ie}{\textit{i.e.},\xspace}

\newcommand*{\etal}{\textit{et~al.}\xspace}

\newcolumntype{L}[1]{>{\raggedright\let\newline\\\arraybackslash\hspace{0pt}}m{#1}}
\newcolumntype{C}[1]{>{\centering\let\newline\\\arraybackslash\hspace{0pt}}m{#1}}
\newcolumntype{R}[1]{>{\raggedleft\let\newline\\\arraybackslash\hspace{0pt}}m{#1}}

\newcommand\pquote[2]{{``\textit{#2}'' (P#1)}}
\newcommand\epquote[2]{{``\textit{#2}'' (PP#1)}}

\makeatletter
\def\thickhline{%
  \noalign{\ifnum0=`}\fi\hrule \@height \thickarrayrulewidth \futurelet
   \reserved@a\@xthickhline}
\def\@xthickhline{\ifx\reserved@a\thickhline
               \vskip\doublerulesep
               \vskip-\thickarrayrulewidth
             \fi
      \ifnum0=`{\fi}}
\makeatother

\makeatletter
\def\thickhlinespace{%
  \addlinespace[1ex]
  \noalign{\ifnum0=`}\fi\hrule \@height \thickarrayrulewidth \futurelet
   \reserved@a\@xthickhline
   \addlinespace[1ex]
   }
\def\@xthickhlinespace{\ifx\reserved@a\thickhline
               \vskip\doublerulesep
               \vskip-\thickarrayrulewidth
             \fi
      \ifnum0=`{\fi}}
\makeatother

\newlength{\thickarrayrulewidth}
\newlength\Origarrayrulewidth

\algnewcommand{\IfThenElse}[3]{
  \State \algorithmicif\ #1\ \algorithmicthen\ #2\ \algorithmicelse\ #3}

\definecolor{downredcolor}{HTML}{e31a1c}
\definecolor{upgreencolor}{HTML}{33a02c}

\definecolor{DarkGreen}{HTML}{5DAC81}
\newcommand\review[1]{\textcolor{black}{#1}}

\newcommand\imwutreview[1]{\textcolor{black}{#1}}

\newcommand\projectname{Talk2Care}

\renewcommand{\shortauthors}{Yang et al.}
\begin{document}


\title{\projectname{}: Facilitating Asynchronous Patient-Provider Communication with Large-Language-Model}

\author{Ziqi Yang}
\email{yangziqi@umich.edu}
\authornote{Both authors contributed equally to this research.}
\affiliation{%
  \institution{University of Michigan}
  \country{USA}
  \postcode{43017-6221}
}

\author{Xuhai Xu}
\authornotemark[1]
\email{xoxu@mit.edu}
\affiliation{%
  \institution{Massachusetts Institute of Technology}
  \country{USA}
}

\author{Bingsheng Yao}
\email{yaob@rpi.edu}
\affiliation{%
  \institution{Rensselaer Polytechnic Institute}
  \country{USA}
}

\author{Shao Zhang}
\email{zhang.shao1@northeastern.edu}
\affiliation{%
  \institution{Northeastern University}
  \country{USA}
}

\author{Ethan Rogers}
\email{rogers.eth@northeastern.edu}
\affiliation{%
  \institution{Northeastern University}
  \country{USA}
}

\author{Stephen Intille}
\email{s.intille@northeastern.edu}
\affiliation{%
  \institution{Northeastern University}
  \country{USA}
}

\author{Nawar Shara}
\email{Nawar.Shara@Medstar.net}
\affiliation{%
  \institution{MedStar Health Research Institute}
  \country{USA}
}

\author{Guodong (Gordon) Gao}
\email{gordon.gao@jhu.edu}
\affiliation{%
  \institution{John Hopkins University}
  \country{USA}
}

\author{Dakuo Wang}
\authornote{Corresponding author d.wang@northeastern.edu}
\email{d.wang@northeastern.edu}
\affiliation{%
  \institution{Northeastern University}
  \country{USA}
}

\renewcommand{\shortauthors}{Yang et al.}
\renewcommand{\shorttitle}{\projectname{}}

\begin{abstract}

Despite the plethora of telehealth applications to assist home-based older adults and healthcare providers, basic messaging and phone calls are still the most common communication methods, which suffer from limited availability, information loss, and process inefficiencies.
One promising solution to facilitate patient-provider communication is to leverage large language models (LLMs) with their powerful natural conversation and summarization capability.
However, there is a limited understanding of LLMs' role during the communication.
We first conducted two interview studies with both older adults (N=10) and healthcare providers (N=9) to understand their needs and opportunities for LLMs in patient-provider asynchronous communication.
Based on the insights, we built an LLM-powered communication system, \projectname{}, and designed interactive components for both groups: (1) For older adults, we leveraged the convenience and accessibility of voice assistants (VAs) and built an LLM-powered \review{conversational} interface for effective information collection.
(2) For health providers, we built an LLM-based dashboard to summarize and present important health information based on older adults' conversations with the VA.
We further conducted two user studies with older adults and providers to evaluate the usability of the system.
The results showed that \projectname{} could facilitate the communication process, enrich the health information collected from older adults, and considerably save providers' efforts and time.
We envision our work as an initial exploration of LLMs' capability in the intersection of healthcare and interpersonal communication.


\end{abstract}




\maketitle
\begin{figure}
    \centering
    \includegraphics[width=1\linewidth]{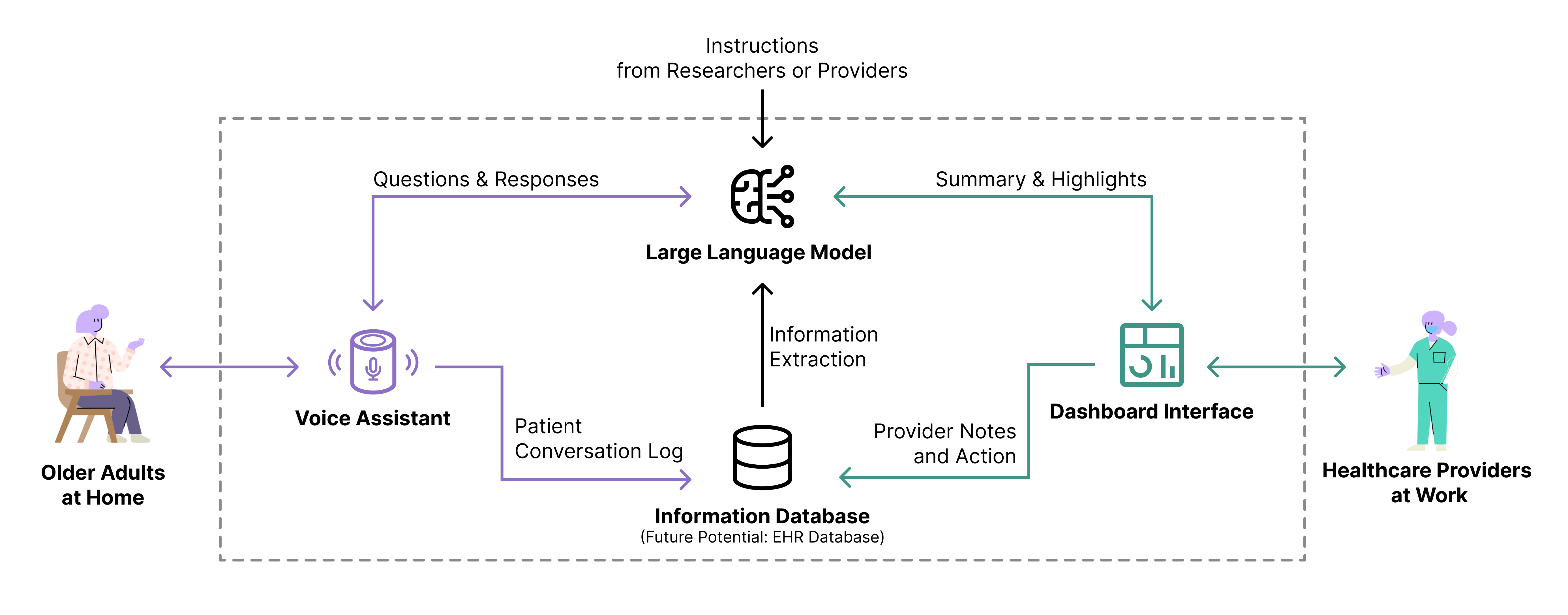}
    \caption{Overview of \projectname{} System. The system consists of two modules. 1) The patient module: An LLM-powered VA interface (in purple) that generates natural conversation with home-based older adults to collect health information and forward it to healthcare providers;
    2) The provider module: A dashboard interface (in green) that summarizes the key information from the older patient conversation to assist providers who are responsible for communication (\eg nurses and patient navigators).
    Note that \projectname{} does not provide specific healthcare advice. Our current implementation does not involve an actual electronic health record (EHR) system, which can be a promising future direction.
    }
    \label{fig:system-overview}
\end{figure}

\section{Introduction}





\review{The healthcare needs of older adults are uniquely critical, with a significant proportion living with chronic diseases that necessitate long-term monitoring and regular interaction with healthcare providers~\cite{NCOA_report}. According to the National Council on Aging (NCOA), nearly 95\% of older adults in the US have at least one chronic health condition, and about 80\% have multiple conditions~\cite{NCOA_report}, amounting to over 50 million seniors needing healthcare support. }
On the one hand, older people often experience a gradual decline in cognitive and physical abilities, which can make regular outpatient visits increasingly challenging. 
On the other hand, healthcare providers are already overloaded and have limited bandwidth~\cite{wang_brilliant_2021,vugs2023clinician}. Asking them to visit patients' homes to check their health is not a scalable solution due to the significant workload~\cite{theile2011home}.
As such, accessible home-based healthcare, which bridges the gap between older adults and their healthcare providers, emerges as one of the promising solutions. It can offer a familiar and convenient environment to older adults while saving healthcare providers' efforts.

\imwutreview{Research has focused on various home-based telehealth applications to support asynchronous communication between older adults and their healthcare providers, such as remote health monitoring  \cite{malasinghe2019remote, halloran2019remote, bates2011wireless, johnson2022remote, crossley2011connect}, healthcare plan management  \cite{arguello_companion_2023}, and patient-provider collaborative review~\cite{chung2016boundary, chung_supporting_2017, schroeder2017supporting}. However, these platforms only support specific data formats and contexts that are tailored to some specific common diseases.}
\imwutreview{When it comes to various daily health communication scenarios (\eg unexpected health concerns), older adults often have to refer to traditional asynchronous communication methods such as a basic text or voice messaging system, as healthcare providers are often not available in real-time for a call (\eg with a busy schedule or outside working hours). However, there are critical gaps in these methods. Firstly, getting feedback and progress via asynchronous messages is time-consuming, and digital messages could be lost due to providers' hasty schedules~\cite{laukka2020health}. Also, older adults may fail to identify crucial symptoms in self-report messages due to limited health literacy. Although questionnaires can be used for more informative communication, they are mostly fixed and lack adaptability for different individual conditions~\cite{kruse2020utilization, park2012effects}. 
Meanwhile, older adults generally have lower technology proficiency compared to younger generations, which is another common obstacle to accessing and using some of these tools~\cite{haase2021older, Smith_2014, kruse2020utilization, foster2014facilitators}. }
For healthcare providers with certain patient follow-up protocols (\eg post-surgery follow-up), making manual phone calls is one of the most common practices, yet it costs a significant amount of workload, and sometimes older adults may not be reachable. 
\imwutreview{The issues in efficiency, adaptability, and accessibility within current communication methods exert communication challenges on older adults and their healthcare providers.}


\imwutreview{A novel conversational interface between older adults at home and healthcare providers may fill in these gaps in asynchronous communication by providing an accessible, interactive, and intelligent communication system.}
The recent technological boost of large language models (LLMs) brings us new opportunities to better achieve the goal.
\imwutreview{The HCI community has explored the application of LLMs for various healthcare practices, such as mental health support~\cite{kumar_2023_exploring,lamichhane2023evaluation}, health information seeking~\cite{yunxiang2023chatdoctor, wei_leveraging_2023,montagna2023data}, and public health interventions~\cite{jo_understanding_2023}, revealing LLMs' intelligence in processing health-related information}. However, there is little previous work leveraging LLMs for patient-provider communication.
In our case, we can leverage its outstanding power of natural language conversation generation and information summarization.
For older adults, an LLM-based \review{conversational agent (CA) }may provide convenience and easy accessibility. \imwutreview{Given older adults' limited health literacy and complicated health conditions, the LLM-powered CA may help older adults better convey their information asynchronously to providers when needed, such as in cases with unexpected health concerns.}
Meanwhile, providers can also use LLMs to automate the information-gathering and summarization process with high-quality data collected from older adults.



However, the role of LLMs in the asynchronous communication between older adults and healthcare providers is still unclear. To address this gap, we first conducted two semi-structured interviews to understand communication challenges and older adults' and providers' perceptions towards LLMs-based communication systems.
Our studies involved both older adults with various health conditions (N=10), as well as health providers (N=9).
Our results revealed a set of asynchronous communication challenges between older adults and healthcare providers.
\imwutreview{We found that it is hard for patients and providers to reach each other, and much information often requires follow-up, explanation, and analysis. These results illustrate the high potential of an LLM-based communication tool.}

Leveraging the insights from our interview study, we designed and implemented a pilot LLM-based system, \projectname{}, to support diverse home-based healthcare communication scenarios.
The system consists of two modules. (1) A patient-facing module: A \review{conversational interface presented as a voice assistant (VA)} that provides a natural conversation experience with older patients to collect health information. (2) A provider-facing module: A dashboard interface for health providers that summarizes and highlights key information from patient-VA conversations. It is worth noting that our system \textit{does not} aim to provide any specific health advice due to the early-stage exploratory nature of this work and ethical concerns.
As discussed later, the ethical concerns must eventually be addressed, but in this work, we explore the value of LLM-driven communication as a first step.
\review{We evaluated the pilot \projectname{} in two cases close to real-world scenarios, one usually initiated by older adults (unexpected health concerns) and one usually initiated by providers (post-surgery follow-up).}
We conducted two user studies with both sides to evaluate the usability of \projectname{}.
\imwutreview{Our results suggest that our LLM-powered communication tool can facilitate the asynchronous communication process, enrich the health information collected from older adults, and save providers' efforts and time considerably. }

Our primary contributions are:
\begin{itemize}
    \item We conducted two interview studies with both older adults and healthcare providers to understand the communication challenges and opportunities for LLMs to assist patient-provider communication.
    \item We designed and developed a pilot LLM-powered system, \projectname{} -- with a VA interface for older patients and a dashboard interface for providers -- to facilitate home-based patient-provider communication for older adults.
    \item We evaluated our systems with two common home-based healthcare scenarios via two user studies with older adults and providers to demonstrate the usability and effectiveness of \projectname{}.
    \item We revealed potentials and design implications for future LLM-powered systems to facilitate and mediate remote patient-provider communication in a human-AI collaborative workflow.
\end{itemize}

\section{Related Work}
\label{sec:related_work}

We first provide an overview of telehealth and patient-provider communication solutions (Section \ref{sub:related_work:challenges}). We then briefly overview intelligent voice assistants for older adults' healthcare (Section \ref{sub:related_work:VA}) and large language models for healthcare (Section \ref{sub:related_work:LLM}).

\subsection{Telehealth and Patient-Provider Communication Solutions}
\label{sub:related_work:challenges}

Telehealth, or telemedicine, refers to the use of digital communication technologies to provide healthcare services remotely \cite{tuckson_telehealth_2017}.
In the past decade, research has shown the advantages of telehealth in saving patients' and providers' time and effort compared to in-person visits~\cite{dorsey2016state, brophy2017overview, crossley2011connect, chen2019home}, especially during the pandemic isolation~\cite{blandford2020opportunities}.
Researchers have explored novel and more accessible tools to enrich the amount of patient health information in presented artifacts for patient-provider collaborative review and remote patient management, such as mobile health applications~\cite{ding_boundary,xu2019leveraging, wang_please_2020,lu2021cultural,xu2022typeout, chung_2019_identify, li_2017_privacy}, sensor systems and instruction platforms~\cite{halloran2019remote, boulanger2013stroke, pandit_exercisecheck_2019, gronvall_beyond_2013, kayser2023layperson,xu2022globem,xu2023globem}, personal coaches ~\cite{arguello_companion_2023}, online platforms for healthcare consultation~\cite{vinker2007web, umefjord2008medical}.
\imwutreview{However, since these applications are usually designed for a specific common disease or developed in an isolated platform with a specific format, they are not adaptable to different patient needs, hard to integrate with existing healthcare systems, and has poor usability~\cite{dorsey2016state, shaver2022state, donelan2019patient, vugs2023clinician, kruse2020utilization, park2012effects}.}
Especially for older adults who have relatively low health and technical literacy, most of the existing solutions still cannot provide adequate accessibility and usability ~\cite{kruse2020utilization, rao_design_2020, Smith_2014, foster2014facilitators}.
Therefore, a great number of older patients, still refer to traditional communication methods such as phone calls~\cite{dorsey2016state, rao_design_2020}.
Similarly, healthcare providers also often find these telehealth tools hard to use~\cite{jawhari2016benefits, kasim2023usability, kawakami2022improving}. \imwutreview{Therefore, making phone calls and leaving messages are still popular communication methods for providers in cases such as hospital discharge follow-up~\cite{mark2022addiction, kasim2023usability}.}
\imwutreview{However, these solutions lacked efficiency and adaptivity. Due to limited provider availability, communicating via asynchronous messages is time-consuming, and digital messages could be lost due to providers' hasty schedules~\cite{laukka2020health}. Also, when the messages rely on older adults' descriptive self-reports, crucial symptoms could be missed due to their limited literacy~\cite{rubin2011associations, chesser2016health}. Another option is to integrate structured questionnaires, yet they are not adaptive or flexible enough for individuals' personalized conditions and needs~\cite{eack2006limitations, madan2012pre}. These limitations in current communication technologies call for an intelligent and accessible solution to facilitate their patient-provider.}
In this paper, we explore a novel, easy-to-use, and natural communication solution for both home-based older adults and healthcare providers.





\subsection{\review{Intelligent Conversational Agents for Older Adults Healthcare}}
\label{sub:related_work:VA}

\review{Researchers have explored a variety of CAs to effectively support older adults' home-based healthcare needs, including chatbots~\cite{ryu_simple_2020, gudala_benefits_2022}, embodied or physical CAs~\cite{simpson_daisy_2020, spillane2018adele, bartle_second_2022, bartle_this_2023}, and intelligent VAs~\cite{pradhan_use_2020, desai_ok_2023, berube_reliability_2021, harrington_its_2022}.  Among them, VAs have stood out with their simplicity, convenience, and good accessibility in scenarios including health information seeking ~\cite{pradhan_use_2020, brewer_empirical_2022, berube_reliability_2021, harrington_its_2022, jain_2018_farmchat}, learning ~\cite{desai_ok_2023}, assisting home health aides ~\cite{bartle_second_2022, bartle_this_2023}, and mental support~\cite{spillane2018adele, stara_usability_2020, simpson_daisy_2020}. Yet studies have also pointed out the limited use of these VAs by older adults due to usability barriers, cultural gaps, lack of explainability, and flexibility ~\cite{harrington_its_2022, trajkova_alexa_2020, gudala_benefits_2022, zubatiy_empowering_2021, blair_understanding_2019}, and none of them have explored CAs as communication interfaces. This motivated us to explore novel CAs to fill in the gaps in older adults' home-based healthcare and communication, taking VA as an example.}
\review{Some recent studies have explored the combination of conversational user interfaces and more intelligent techniques. }
For example, Natural Language Processing (NLP) methods have been used to build \review{CAs} in travel apps and browsers to simplify user interaction~\cite{poran2022with, cambre_firefox_2021}\review{, and LLMs have been used in CAs to interact with mobile UI and programming tasks~\cite{wang_2023_enabling, ross_programmers_2023}}.
Recently, Zhang \etal proposed SpeechGPT, an LLM with intrinsic cross-modal conversational abilities~\cite{zhang_speechgpt_2023}.
Rubenstein \etal presented AudioPaLM, a unified speech-text LLM~\cite{rubenstein2023audiopalm}. However, little is known about whether these could be applied to real-world contexts, especially for home-based older adults.
Our work seeks to leverage this interaction paradigm to build an interactive and easy-to-use LLM-powered \review{CAs that advances} personalized information gathering and asynchronous communication with healthcare providers. 



\subsection{Large Language Models for Healthcare}
\label{sub:related_work:LLM}

Recently, the technology boost of Large Language Models (LLM), such as ChatGPT~\cite{OpenAI_2022}, has shown its great potential in engaging in, scaffolding, and processing natural conversations~\cite{xiao2020tell, xiao2023supporting,shen2023convxai, hamalainen_evaluating_2023, wang_2023_enabling, liu2023human}.
The latest studies have discussed and explored its application for individual patients' in health information seeking~\cite{yunxiang2023chatdoctor, xiao2023powering}, mental health support~\cite{lamichhane2023evaluation, kumar_2023_exploring,xu2023mentalllm}, personal health coaching and management~\cite{wei_leveraging_2023, montagna2023data} and health education~\cite{montagna2023data, kung2023performance}. 
For healthcare professionals, researchers have also leveraged LLMs to support public health interventions ~\cite{jo_understanding_2023}, clinical pre-screening~\cite{hamer_improving_2023, wang_brilliant_2021}, risk prediction ~\cite{o2015risk,gatt2022review, kennedy2014delirium, beede2020human} and information processing~\cite{kocaballi_envisioning_2020, nair_generating_2023,  cascella2023evaluating, nori2023capabilities, agrawal_large_2022, arbatti_what_2023}. 
However, multiple studies have raised concerns about using LLMs for healthcare scenarios, due to their inconsistency, potential errors, and bias in clinical work~\cite{kocaballi_envisioning_2020, iliashenko2019opportunities, berube_reliability_2021}.
Due to these concerns, in our work, we focus on facilitating communication, but not on providing any specific healthcare advice.

Despite the great potential of LLM for promoting healthcare work and experiences, little work aims to address the challenges of patient-provider communication mentioned in Section~\ref{sub:related_work:challenges}, and little is know about how LLM might help patient-provider communication and benefit both stakeholders.
Close to our research, Jo \etal conducted interviews with stakeholders in an LLM-powered system that supports public mental health interventions for people living alone~\cite{jo_understanding_2023}.
The study revealed that the system could reduce the workload for mental health practitioners \review{while providing} mental support for end users in large-scale healthcare services.
However, the existing solution still failed to meet the personalized requirements of users at home because it followed a uniform setup(``one-size-fits-all''), which lacked the necessary flexibility for customization~\cite{jo_understanding_2023}.
\imwutreview{With our pilot system, we aim to facilitate patient-provider communication by providing older adults with more interactive, responsive, and adaptive experience, and providing healthcare providers with more efficient information analysis assistance.}

\section{Need-finding Study}
\label{sec:understand}

\imwutreview{To better understand the communication gap between older adults and their healthcare providers and to identify user needs, we conducted two semi-structured interview studies, one with older adults (Section~\ref{sub:understand:older_adults}), and the other with healthcare providers (Section~\ref{sub:understand:providers}).}
\imwutreview{Through our interview studies, we identify (1) communication challenges such as limited provider availability and disparities in literacy, (2) the opportunities for future LLM-powered systems to offload providers and provide mental support, as well as (3) reliability and ethical concerns.
Understanding these challenges and user needs suggests the potential of LLM-powered systems to facilitate communication. Moreover, our results guide the design of our system in Section~\ref{sec:system}.}

\subsection{\imwutreview{Interview Study 1: Older Adults}}

\label{sub:understand:older_adults}
\subsubsection{\imwutreview{Methods: Participants and Procedure}}
\label{subsub:understand:older_adults:participants}
After obtaining the IRB approval, we recruited older adults (N=10) from a local university-maintained participant pool of older adults with snowball sampling.
\review{Participants lived in an area in the northeast part of the United States.}
All the older adults resided in single-family homes and were 65 years of age or older.
Participants reported one or more chronic conditions and urgent care histories; they had rich experience communicating with their healthcare providers in recent years. Out of the ten older adult participants, two of them had experience using AI  \review{  (\eg using ChatGPT on its web interface)}, six knew about the concept but had not used AI, and two others were unfamiliar with the concept.
Table~\ref{tab:study1_demo} summarizes their detailed demographic information\footnote{We show the age of participants by age range for deidentification purposes.}.

\renewcommand{\arraystretch}{1.3}
\begin{table}[h]
    \centering
    \resizebox{1\linewidth}{!}{
    \begin{tabular}{c|c|c|l|l}
    \hline \hline
        \textbf{P\#} & \textbf{Age} & \textbf{Gender} & \textbf{Chronic/Occasional Condition} & \textbf{Surgery/Accident History} \\
        \hline
        P1 & 65-69 & Female  & Migraine, hypertension, former depression, dementia risk & Joint surgery \\ 
        P2 & 75-79& Female  & Hypertension, arthritis, stomach issue, anxiety disorder & Thumb injury \\ 
        P3 & 65-69 & Female  & Generally healthy, sinus infection, cervical dystonia & / \\ 
        P4 & 80-84 & Female & Generally healthy, former alopecia & Ankle injury \\ 
        P5 & 65-69  & Female & Generally healthy & Throat \& shoulder surgery, body injury \\ 
        P6 & 80-84 & Female  & Atrial fibrillation, occasional pain& / \\ 
        P7 & 70-74 & Female  & Generally healthy, usual aches and pain & / \\ 
        P8 & 70-74  & Male  & COPD, sleep apnea, atrial fibrillation& / \\ 
        P9 & 65-69 & Male  & Generally healthy, long-term COVID influence & Collarbone injury\\ 
        P10 & 70-74  & Female  & Arthritics, occasional depression, pain and hypertension & / \\ 
        \hline \hline
    \end{tabular}
    }
    \caption{Demographics of Older Adult Participants}
    \label{tab:study1_demo}
\end{table}
\renewcommand{\arraystretch}{1.0}

\label{subsub:understand:older_adults:design}
Our interview sessions followed a semi-structured format.
Our key interview questions covered three perspectives:
(1) What were older adults' needs and challenges in communicating with their healthcare providers at home?
(2) How might an LLM-based \review{CA} system facilitate communication and address these needs and pain points?
(3) What concerns and risks would they have when applying such an AI-powered system?
\review{For the participants who had no experience interacting with LLMs, we used slides to introduce the concept of AI and LLM in descriptive words and visual cues and presented an example conversation with ChatGPT on the web interface. }Based on these starting points, the interviewer followed up with interviewees based on their responses. The specific questions can be found in Appendix \ref{sub:appendix-script-older} and Appendix \ref{sub:appendix-script-provider}.
The study lasted 25-30 minutes. Participants were compensated \$25 for their time.
The interviews were recorded via Zoom and transcribed afterward.
Thematic analysis was employed to analyze the data. Two authors coded interview transcripts iteratively until they reached a consensus. \imwutreview{We summarize our findings as follows.}

\subsubsection{\imwutreview{Findings: Communication Needs and Challenges}}
\label{subsub:understand:older_adults:results_challenges}
Participants reported a variety of examples where they needed to communicate directly with their healthcare providers at home, such as daily questions about new symptoms and urgent care (\eg COVID or flu symptoms), confusion or changes about an existing condition (\eg occasional pain or blood pressure change, prescription instruction), and post-surgery-related questions. Many cases were consistent with, and supported by, prior research~\cite{gronvall_beyond_2013, vugs2023clinician, donelan2019patient, mitzner2009older}. 
Most of these communication needs were case-specific and personalized. Participants' current solutions included visiting/calling healthcare services, communicating through the patient portal with providers, discussing with friends or family, and seeking online resources as alternatives.
Our interview revealed a set of challenges \review{to} these existing solutions.

\imwutreview{\textbf{Communication Inefficiency}}
Despite the wish to get quick and in-time responses from the healthcare system and providers, 5 participants mentioned that it was \textbf{hard to get to healthcare providers} because they were often too busy during working hours.
\imwutreview{For example, \pquote{10}{
then you're on hold for maybe 30 minutes, which is crazy.}}
The in-person visiting option posed more challenges. Besides the traveling difficulties, the waiting time for an appointment with providers often took several weeks or even months.

Even in cases where participants got to talk to providers, some information was often ambiguous or even \imwutreview{\textbf{lost during communication}}. For example, P1 mentioned that nurse practitioners ``\textit{never put down exactly what I say... They may not realize that it is very significant}''.
P8 gave another example where the messages from providers and lab results \review{often lacked} explanations and \review{were} hard to understand.
Such an information loss caused the participant to pay extra effort to describe things or request explanations one more time when they talked to doctors \review{in subsequent interactions}, which reduced communication efficiency and brought confusion for older adults.

Some participants were more familiar with technology and learned to use digital patient communication portals to leave messages to providers, which could address these challenges to some extent. However, \imwutreview{\textbf{lack of digital literacy}} is a widely known issue~\cite{pradhan_use_2020, Smith_2014, foster2014facilitators}. 
\imwutreview{Only four out of the ten participants knew how to use such digital technology.}


\textbf{Reliability Concerns of Alternative Resource.}
Since they could not get in-time communication with health providers, \review{four} participants reported resorting to alternative resources by searching online or talking to families.
Consistent with previous work~\cite{brewer_empirical_2022, berube_reliability_2021}, they expressed concerns about the reliability of these resources. \pquote{1}{I'd rather just go to the doctor. That's their area of expertise.}
This became another strong motivation for us to facilitate communication between older adults and providers.

\subsubsection{\imwutreview{Findings: Opportunities for An LLM-facilitated CA System}}
\label{subsub:understand:older_adults:results_LLM}
To further understand how LLMs may help older adults with their communication, we introduced the concept of an LLM-based  \review{CA with the possibility of implementing a VA as the user interface}. We also presented examples of how such a system could have natural conversations, collect information and conditions, and forward the information to healthcare providers for further decision and contact. We highlighted explicitly that our system would not provide any specific health advice, but the system could provide general information about clinics, procedures, or term explanations.
Overall, older participants expressed strong interest in such an AI-powered system to help them address the communication challenges. 

\textbf{Offloading for Healthcare Providers.}
As introduced in Section~\ref{subsub:understand:older_adults:results_challenges}, one of the key communication challenges was the unavailability of healthcare providers. Older adults hoped that the system could help save providers' efforts. 
\imwutreview{\pquote{8}{... Whereas with a computer, and assuming it can respond to 100 people at the same time ...}}
\imwutreview{This could potentially help providers with tedious procedure work and give them the bandwidth to respond to patients' actual inquiries.}

\imwutreview{\textbf{Emotional and Accessibility Support}}
\imwutreview{Although we made it clear that the system would not provide specific health advice, 3 participants still thought it could give immediate responses and provide \textbf{emotional support}}.
\imwutreview{For example, P4 thought the system could accompany her at home since she lives alone. P10 also agreed that the audio modality of the VA would be comforting, \pquote{10}{Sometimes you just wanted to ... hear a voice}.}
Moreover, the support could also be expanded to other family members. For example, P1 shared her post-surgery experience and thought that if such a system existed, it would \pquote{1}{give my daughter enough comfort to let me be in our apartment alone after}.

\imwutreview{In addition, participants commented that the natural VA system could address the technology barrier of navigating digital communication portals by improving cognitive accessibility the intelligence from LLMs.}
\imwutreview{For example, P1 had a potential dementia risk and said ``\textit{there's going to come a time when I'm not going to be able to remember how to use it.}'' }

\textbf{Function Expectations.}
\imwutreview{In addition to the natural conversation, information summary, and forwarding features, participants also suggested other expectations, including having pre-set personalized instructions from doctors (P8), providing high-level suggestions and references (P1, P6), and delivering administrative information (P2, P10). }

\subsubsection{\imwutreview{Findings: Concerns and Risks}}
\label{subsub:understand:older_adults:results_concerns}
We also investigated the potential concerns of older adults when using the intelligent system.
We found the common ethical concerns for AI systems~\cite{zhuo2023exploring} and privacy concerns of VAs from 3 participants~\cite{pradhan_use_2020}.
For example, P5 was concerned that they would accidentally reveal sensitive personal information: ``\textit{because it's a little bit [listening to] everything}''. However, many participants did not have such concerns, as long as it ``\textit{follows the criteria that the federal regulations}'' (P2).
Besides, some participants were also concerned about the learning curve of the system, which we will evaluate through a user study in Section~\ref{sub:evaluation:older_adults}.

\subsection{\imwutreview{Interview Study 2: Healthcare Providers}}
\label{sub:understand:providers}


\subsubsection{\imwutreview{Methods: Participants and Procedure}}
\label{subsub:understand:providers:participants}
By emails and snowball sampling, we recruited 9 nurse practitioners, doctors who are primary care physicians or specialists, and other professionals who have close contact with patients. 
Participants have frequent communication with patients at home, and most of them have close communication with older adult patients. 
Six of the nine participants had experience with AI \review{  (\eg using ChatGPT on its web interface)}, and three others knew about the concept.
Table \ref{tab:study2_demo} summarizes their demographics and expertise.
\imwutreview{We mainly recruited physicians because we focus on the communication needs concerning clinical decisions or action in this study. In these scenarios crucial to older adults' health, physicians are the ones who are mainly responsible, while nurse practitioners or other staff largely worked on administrative work. In addition, the older adult participants expressed their strong wish to closely communicate with physicians, as discussed in Section \ref{subsub:understand:older_adults:results_challenges}  We include providers with different roles to gain a holistic view of the providers' perspectives.}

\renewcommand{\arraystretch}{1.3}
\begin{table}[ht]
    \centering
    \resizebox{1\linewidth}{!}{
    \begin{tabular}{c|c|l|c}
    \hline \hline
        \textbf{PP\#} & \textbf{Gender} & \textbf{Role, Responsibility, Specialty} & \makecell{\textbf{Years of Experience}\\\textbf{in Healthcare}} \\ \hline
        PP1 & Female & Primary care physician, specialist in diabetes & 10-20 Years \\ 
        PP2 & Female & Nurse practitioner in emergency department & 0-2 Years \\ 
        PP3 & Female & Dentist, health consultant & 0-2 Years \\ 
        PP4 & Male & Palliative medicine physician & >20 Years \\ 
        PP5 & Male & Primary care physician, internist, pediatrician, hospice and palliative care specialist& >20 Years \\ 
        PP6 & Male & Internal medicine physician & 2-5 Years \\ 
        PP7 & Female & Clinical research assistant \& medical records coordinator & 5-10 Years \\ 
        PP8 & Female & Palliative care nurse practitioner, former patient navigator & >20 Years \\ 
        PP9 & Female & Pediatrician, physician in ER & >20 Years  \\ 
        \hline \hline
    \end{tabular}
    }
    \caption{Demographics of Healthcare Provider Participants}
    \label{tab:study2_demo}
\end{table}
\renewcommand{\arraystretch}{1.0}

\label{subsub:understand:providers:design}
We designed a semi-structured interview process, with a focus on three perspectives:
(1) What were healthcare providers' needs and challenges in communicating with their patients at home, especially with older adults?
(2) How might an LLM-based system facilitate their communication with patients and address these needs and challenges?
(3) What concerns and risks would they have when applying such a system?
The specific questions can be found in Appendix \ref{sub:appendix-script-provider}. The interviewer followed up with the providers based on their responses. 
The study lasted 25-30 minutes. Providers were compensated \$25 for their time.
Similar to our procedure in Section \ref{sub:understand:older_adults}, we recorded the Zoom interview, and two researchers used thematic analysis on the transcribed data. We summarize our findings in the following sections from the three perspectives listed above.

\subsubsection{\imwutreview{Findings: Communication Needs and Challenges}}
\label{subsub:understand:providers:results_challenges}
Providers mentioned a variety of scenarios that required patient-provider communication. The majority were about older patients' daily care inquiries and the follow-up of previous treatments or surgeries. Sometimes long-term patient education also requires communication.
Providers did use digital tools to manage information and reminders, yet for older adults, phone calls were still the most common method.
Our interview results suggest several prominent challenges.

\imwutreview{\textbf{Ineffective Communication with Existing Asynchronous Methods.}}
In Section~\ref{subsub:understand:older_adults:results_challenges}, participants mentioned that providers were hard to reach. However, interestingly, providers felt the same.
Four providers found patients, especially older adults, were hard to reach via phone calls. They often needed to call multiple times to finally reach patients, and sometimes they just could not succeed.
Meanwhile, providers mentioned that they are ``understaffed and overworked'' since they usually have too many patients to handle (PP8). Their bandwidth was limited, so they could not spend all their time reaching out to older patients or going through a standardized in-person visit. As a result, providers had to prioritize other tasks, and communication often \review{experienced} inevitable delays.

Existing asynchronous solutions have been trying to address this limited availability challenge. Tools such as patient portals or EHR systems are set up as an alternative to manual phone calls.
\imwutreview{However, providers commented that the messages from patients were usually \textbf{unclear}.} These messages were usually too limited or overly long for providers to make the right decision.
Meanwhile, the design of current EHR systems has too much redundant information for providers to process
\epquote{5}{There's so much information that you need to disregard.}
This may explain the information loss issue mentioned by older adults in Section~\ref{subsub:understand:older_adults:results_challenges}.


\textbf{Extra Effort Communicating with Older Adults.}
Out of the nine providers, seven also mentioned the challenges during the real-time communication with older adults:
Older adults often had limited health literacy and higher accessibility needs. Four of the providers felt that they needed to pay extra attention to language and spend more time explaining concepts to make sure that the information was delivered to and understood by older adults.
\imwutreview{\epquote{2}{... like talking [in a] slower pace or using non-medical terms that any patient can understand.}}


\subsubsection{\imwutreview{Findings: Opportunities for An LLM-facilitated Communication System}}
\label{subsub:understand:providers:results_LLM}
To investigate how LLM-based AI can be used to help providers address communication challenges, we introduce the concept of LLM and our system.
We briefly introduced the concept of both the patient-facing VA system for older adults and the LLM-powered information dashboard that summarizes the health information of older adults. Similar to Study 1, we also explicitly mentioned that the AI system would not provide any specific health advice.
Overall, providers were excited about the potential of the system after learning the concept. We summarize the opportunities below.

\imwutreview{\textbf{Promoting Efficiency}}
In Section~\ref{subsub:understand:older_adults:results_LLM}, older adults expressed the hope that such an LLM-powered system could help offload providers. This was confirmed by our Study 2.
\imwutreview{Providers agreed that using a VA system \& an information dashboard could significantly \textbf{save their effort}.}
\epquote{1}{[The system] could be very helpful for those local community clinics that may cover tens or even hundreds of patients.}
Providers could leverage such an intelligent communication system to reduce the procedural process and focus on more critical tasks.

\imwutreview{Moreover, providers also mentioned that such a system could help to \textbf{address the information loss problem} (Section~\ref{subsub:understand:older_adults:results_challenges} and \ref{subsub:understand:providers:results_challenges}) by recording things and provide analysis results.}
\epquote{7}{Sometimes we weren't able to capture the full experience... AI will come in if it's being recorded... and capture that data.}
Providers would like to see a concise summary of the conversation as they saw during our study (PP1, PP2, PP3), which could help them with the triage and prioritization of the right next step.
These comments guided the design of our dashboard system in Section~\ref{sub:system:providers}.


\textbf{Experience Improvement for Older Adults.}
Consistent with the comments from older adults, providers, from their own perspective, also agreed that the VA system could help with the conversation experience for older adults.
\epquote{7}{[The VA system] might eliminate that feeling of judgment, and the patient will probably answer more accurately, as opposed to sitting across from another person who could potentially, you know, unintentionally making them feel uncomfortable because of the certain questions and answers that are required.}
Providers also agreed with the accessibility advantage of a VA system over text-based solutions \imwutreview{, which strengthens our motivation for the VA system for older adults}. 



\textbf{Function Expectation.}
Providers also brought up a few features that the communication system would be good to have.
For the VA system, several providers mentioned the necessity to provide ``closed loop communication''. \epquote{2}{That would be helpful and increase the accuracy of the information.}
Meanwhile, providers also mentioned the expectation of such an AI system to be integrated with EHR systems (PP1, PP4).
Some providers also brought up the importance of recording conversation audio to assist with certain diagnoses.

\subsubsection{\imwutreview{Findings: Concerns and Risks}}
\label{subsub:understand:providers:results_concerns}
Providers had a strong agreement with our design of not providing health advice due to the concern about AI reliability.
Moreover, they also expressed concerns about the risks of misinterpretation (PP2, PP5), AI-embedded bias (PP3, PP7), and sensitive/triggering topics, such as cancers or sex crimes (PP4).
Meanwhile, some were also worried about how AI might influence patient-provider interaction in the long run (PP4, PP9).
Some of these ethical concerns were beyond the scope of this paper, and we will discuss potential solutions in Section~\ref{sec:discussion}.

\subsection{\imwutreview{Findings Summary}}
\label{sub:formative_findings_summary}
\imwutreview{
We summarize our key findings of this section as follows:
\begin{enumerate}
    \item Problems in current communication methods, including provider availability, limited information communicated, and accessibility needs, have led to communication challenges for older adults and their providers.
    \item Participants are highly interested in an LLM-powered system to offload healthcare providers, provide mental support, and gather more information.
    \item Reliability and privacy concerns were raised by both patients and providers.
\end{enumerate}
These insights can guide the design of our LLM-powered system in the next section to facilitate patient-provider communication for older adults. 
}

\section{\projectname{} System}
\label{sec:system}
The insights from older patients and health providers shed light on the design and implementation of our pilot \projectname{} system to facilitate patient-provider communication.
Figure~\ref{fig:system-overview} presents an overview of our pilot system.
\projectname{} consists of two parts:
First, the patient component: an LLM-powered VA interface that asks and answers older patients' questions to collect their health information, which will be forwarded to healthcare providers for future decisions and contact (Section~\ref{sub:system:patients}).
Second, the provider component: a dashboard interface that employs LLM to summarize patients-VA conversations and present important information while keeping original details to save effort and time for healthcare providers (Section~\ref{sub:system:providers}).

\subsection{Patient Component: LLM-powered Voice Assistant}
\label{sub:system:patients}

\begin{figure}
    \centering
    \includegraphics[width=\linewidth]{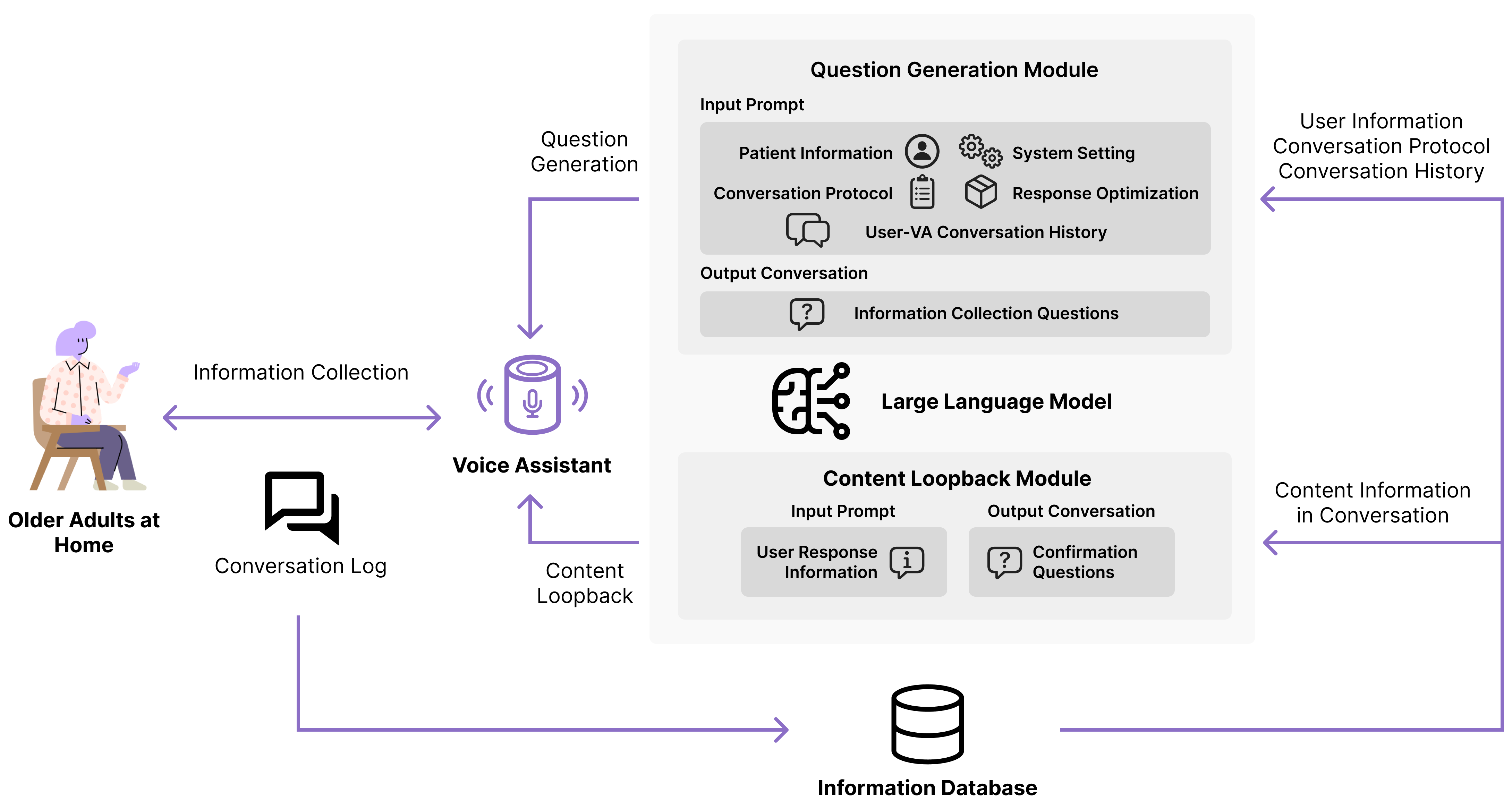}
    \caption{The Component of \projectname{} System for Home-based Older Adults.
    The VA interface has multi-turn personalized conversations with the older adult to collect related health information.
    The LLM-powered Question Generation Module is responsible for taking the older adult's words and generating questions for effective information collection. The prompt design of this module is detailed in Figure~\ref{fig:prompt-design-older-adults}.
    Another LLM-powered Content Loopback Module is to make sure that key information from the older adult (e.g., pain level) is accurate by double-checking the content, a common healthcare communication practice.
    The older adult's information, conversation protocol, and conversation log are stored in the information database.
    }
    \label{fig:system-older-adults}
\end{figure}

The results in Section~\ref{subsub:understand:older_adults:results_LLM} confirmed the advantage of an LLM-powered VA system for its convenience and accessibility for home-based older patients.
We build a VA interface to collect information health information (\eg detailed symptom descriptions). The interface has a multi-turn conversation ability so that they can have continuous and natural back-and-forth conversations with older patients, generating appropriate and personalized follow-up questions to collect important information. 
Figure~\ref{fig:system-older-adults} provides an overview of this component of \projectname{} system, including the two modules for conversation generation (Section~\ref{subsub:system:patients:question} and \ref{subsub:system:patients:loopback}), the VA hardware (Section~\ref{subsub:system:patients:hardware}), and the database (Section~\ref{subsub:system:patients:database}).

\subsubsection{Question Generation Module}
\label{subsub:system:patients:question}
The core part of the VA system is to generate high-quality questions to collect key health information from older adults, which can then be forwarded to healthcare providers for decision-making.
To achieve this goal, we leverage GPT-3.5-Turbo over GPT-4.0 due to the fast speed of GPT-3.5.
LLMs are known to generate offensive or biased texts, sometimes even harmful content~\cite{zhuo2023exploring,hartvigsen2022toxigen}.
To address the ethical concerns,
we conduct a series of prompt designs to improve the question generation referring to prompt optimization practices from previous works~\cite{wei_leveraging_2023, Prompt_2023}.
Specifically, one researcher conducted a set of preliminary prompt designs and tested them with GPT-3.5. Then, another two researchers, one with technical expertise and the other with healthcare expertise, acted as quality checkers for the content and provided feedback. This process was iterated through multiple rounds until all researchers were satisfied with the generated content. \footnote{Despite these efforts, we acknowledge that this could not address the ethical concerns completely. We show more discussion in Section~\ref{sec:discussion}.}

\begin{figure}
    \centering
    \includegraphics[width=\linewidth]{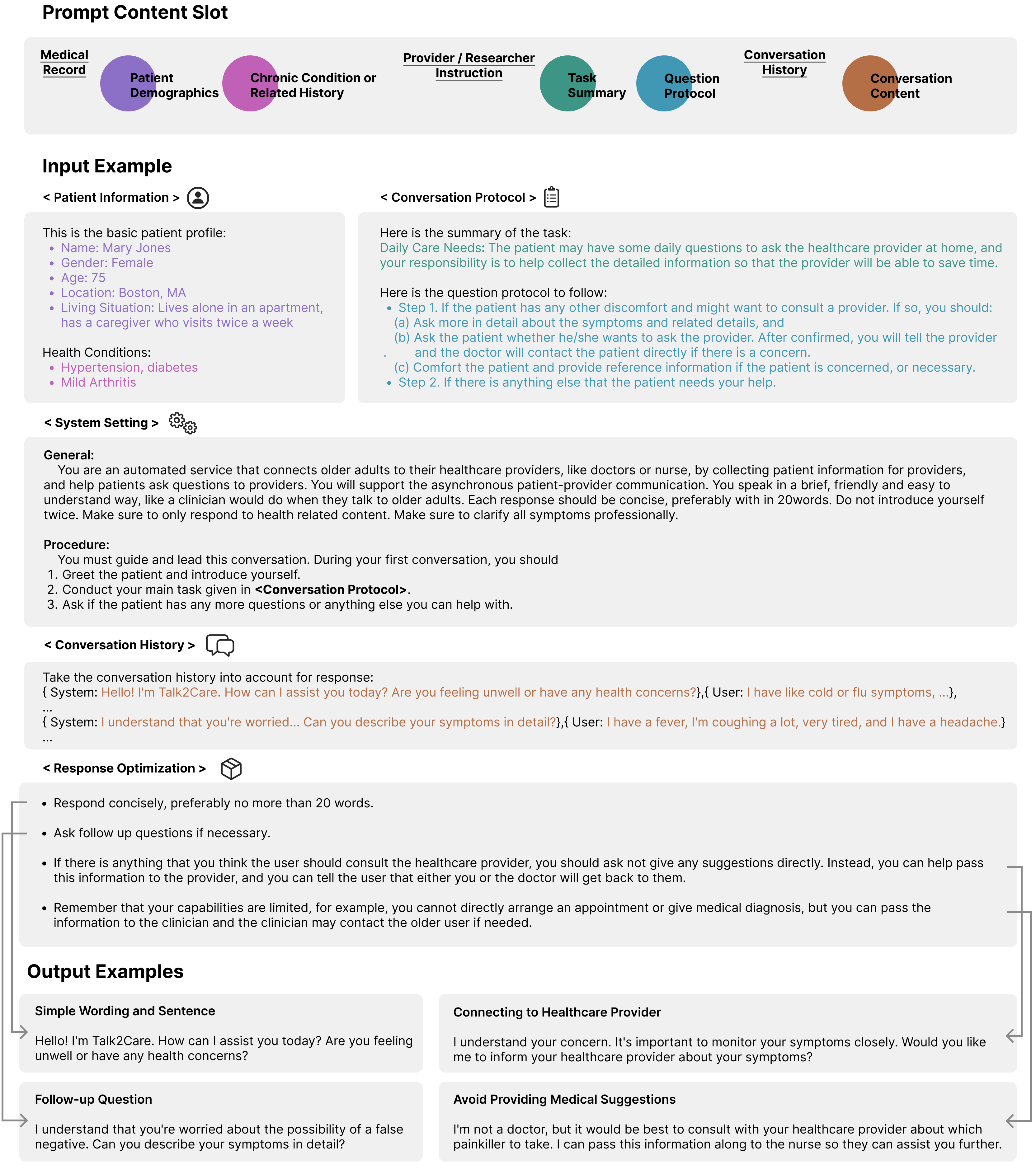}
    \caption{Prompt Design of High-Quality Question Generation for Health Information Collection. The input prompt consists of five parts: 1) patient information, 2) conversation protocol, 3) system setting, 4) conversation history, and 5) response optimization. For multi-turn conversation, 5) will be repeated for each round of conversation.
    The colored texts are parameters that can be extracted from the information database (see Figure~\ref{fig:system-older-adults}).
    Note that the conversation protocol needs to be set by researchers or healthcare providers to ensure question validity. This figure shows an example of daily-care protocol.}
    \label{fig:prompt-design-older-adults}
\end{figure}

Finally, we propose a set of five important content factors to be integrated as a complete prompt, as illustrated in Figure \ref{fig:prompt-design-older-adults}.
\textbf{(1) Patient Information.}
This part includes the older adult's basic information, such as their name, gender, age, and living situation. Moreover, it also includes their health situations such as chronic conditions and medical history. This information was added to the database during the initial setup of the system. 
\textbf{(2) Conversation Protocol.}
To ensure the validity of the information collection questions, the VA should follow a standard protocol employed by providers for a specific task (e.g., daily healthcare needs in Figure~\ref{fig:prompt-design-older-adults}).
Different scenarios may require different protocols (e.g., post-survey follow-up vs. daily care needs).
Therefore, the information database stores a set of protocols collected from online resources, researchers, and providers.
We separate the task summary and question protocol into two chunks to provide a flexible mapping between tasks and protocols.
\footnote{Later in our user study in Section~\ref{sec:evaluation}, we pre-determined the protocol depending on the scenarios. Future work can explore automatic protocol selection depending on the task, as we will discuss in Section~\ref{sec:discussion}.}
\textbf{(3) System Setting.}
This part instructs the role of the VA (as a communication facilitator), its responsibilities (i.e., collecting health information but not giving specific health advice), and its communication style (i.e., concise, health-focused, and easy-to-understand, as suggested in Section~\ref{subsub:understand:providers:results_challenges}). The specific wording of our prompts can be found in Figure~\ref{fig:prompt-design-older-adults}.
\textbf{(4) Conversation History.}
This part includes the conversation history between the older adult and the VA, which can augment the VA's memory and personalize the conversation experience.
\textbf{(5) Response Optimization.}
To optimize the VA's response to the older adult, this part is designed to improve the quality of each response and question generated by GPT-3.5.
For each conversation session, (1) - (4) are only used at the beginning of the generation, while (5) is added in the prompt across all conversation rounds.

\subsubsection{Content Loopback Module}
\label{subsub:system:patients:loopback}
In addition to the question generation, our interviews with providers (Section~\ref{sub:understand:providers}) suggest the importance of content confirmation, especially for important health-related information.
Therefore, we design a simple content loopback module. When a question asks about specific values (e.g., the pain level), the module loops the value back to users and asks them a second time to confirm.

\subsubsection{Voice Assistant Hardware}
\label{subsub:system:patients:hardware}
We experimented with multiple audio devices, including Raspberry Pi and Arduino boards. Finally, we decided to deploy our system on the Amazon Alexa platform by wrapping it as an Alexa Skill, so that we can leverage the well-integrated ecosystem. We use Alexa Echo Dot as the hardware. 
Voice-to-text and text-to-voice service provided by Amazon Alexa is employed to transcribe the older adult's input into texts and convert LLM-generated texts into audio, bridging the text content and natural speech interaction.

\subsubsection{Information Database}
\label{subsub:system:patients:database}
We employ a cloud-based database to store the older adult's information, conversational protocols, as well as conversation log.
These contents are used to construct the prompt for both the question generation module (1, 2, and 4 in Section~\ref{subsub:system:patients:question}) and the content loopback module.
The database is encrypted to protect each individual's privacy.

\subsubsection{Conversation Flow}
Depending on the scenarios, the conversation can be initiated either by older patients (\eg with specific health concerns and care needs) or by the VA (\eg health condition follow-up requests).
In each session, the VA will lead the conversation to ask the patient about symptoms and iteratively gather detailed information to pass to healthcare providers following the prompt instructions.
For example, in the post-surgery scenario, the VA will first ask the patient about the overall health conditions and any pain or discomfort.
If the patient mentions any pain or discomfort, the VA will ask about details (with loopback confirmation), such as pain level, positions of discomfort, and any actions that may influence the symptoms.
If the patient has any confusion about the information, the VA explains the concepts accordingly but does not provide direct medical advice.

\subsubsection{Exception Handling}
We designed the system and prompts to handle common exceptions in voice recognition and user input to minimize unsatisfactory user experience in abrupt failure or termination. 
We use Alexa skill re-prompts~\cite{Alexa_2021} and conversation history to handle unexpected pauses, so that users could have continuous conversations with the VA after pauses.
If voice recognition has errors, users could interrupt the VA and start again.

Some users might seek immediate medical advice from the system, and according to prompts, the system generates responses such as
``\textit{I'm not a doctor, but it would be best to consult with your healthcare provider about which painkiller to take. I can pass this information along to the nurse so they can assist you further. }''(VA)
\imwutreview{Since we aim to take advantage of LLMs, we allow the LLM to provide reasonable explanations for general information assistance, such as ``What is hospice and palliative care?''}
The strategies appeared to be successful in our user study in Section~\ref{sub:evaluation:older_adults}.



\subsection{Provider Component: LLM-powered Information Dashboard}
\label{sub:system:providers}

\begin{figure}
    \centering
    \includegraphics[width=\linewidth]{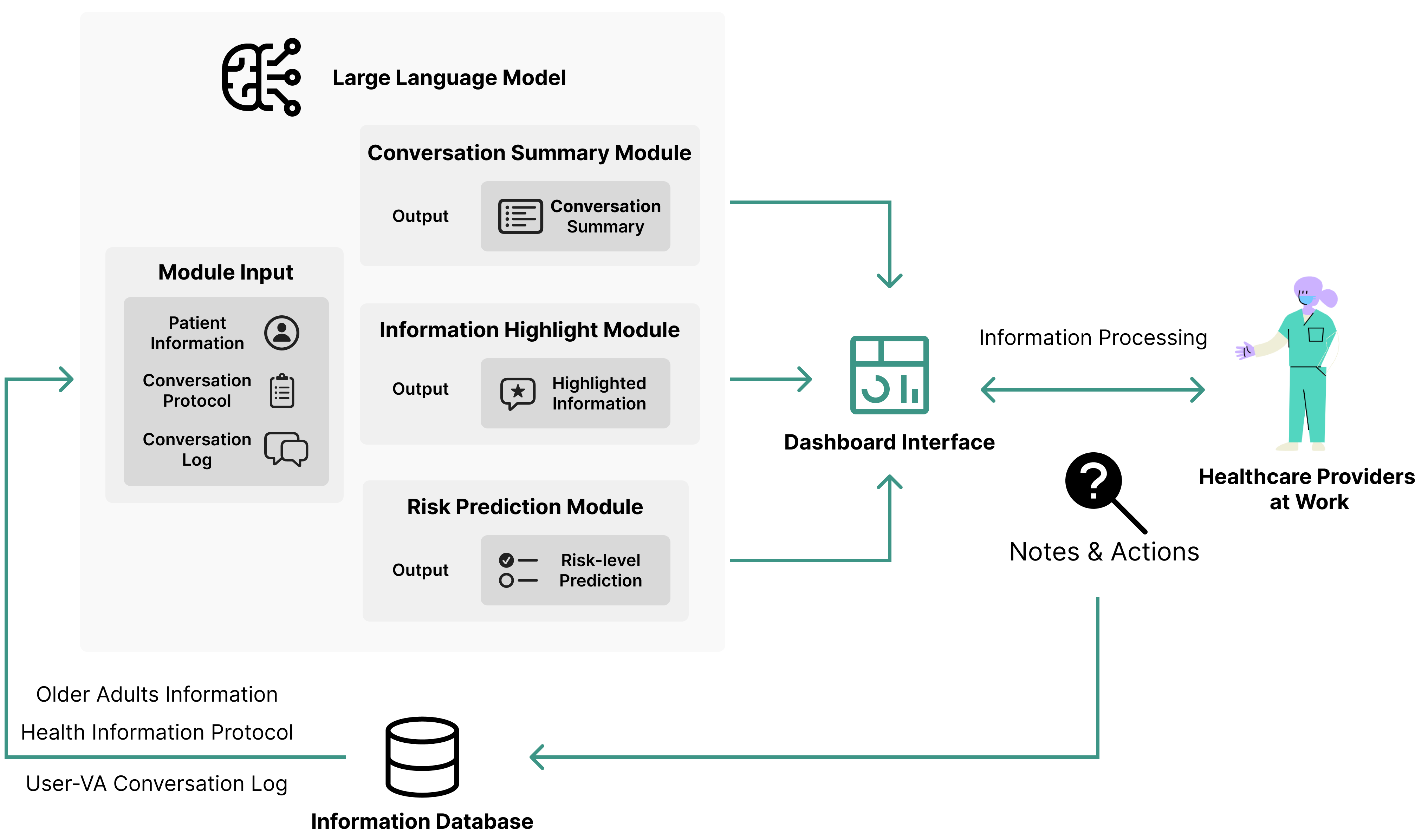}
    \caption{The Component of \projectname{} System for Healthcare Providers. The information dashboard summarizes and highlights key older adults' information.
    The main content on the dashboard is generated by three LLM-powered modules: (1) The Content Summary Module formats the conversation log and user information into a clinical note structure. (2) The Information Highlight Module color-codes the parts in the conversation log that require attention. (3) The Risk Prediction Module suggests the health risk (low, moderate, and high) based on the current conversation log.
    Providers can take notes or further actions on the dashboard, which are then stored in the information database.
    }
    \label{fig:system-healthcare-providers}
\end{figure}

Other than the LLM-powered VA system for older adults, we further developed a dashboard interface for healthcare providers to review the information in the patient-VA conversation and take further actions if necessary.
Figure~\ref{fig:system-healthcare-providers} illustrates an overview of this component of \projectname{}, including three modules for information summary (Section~\ref{subsub:system:providers:summary}), context highlight \ref{subsub:system:providers:highlight}), and risk prediction (Section~\ref{subsub:system:providers:prediction}), together with the dashboard (Section~\ref{subsub:system:providers:interface}) and the database (Section~\ref{subsub:system:providers:database}).

\subsubsection{Conversation Summary Module}
\label{subsub:system:providers:summary}
Our interview results with experts suggest that providing a summary of the conversation between the older adult and the VA system is very helpful (Section~\ref{subsub:understand:providers:results_LLM}).
This module aims to extract the key messages from the conversation log and format them into a clinical notes structure.
To achieve this goal, we also leverage GPT-3.5 with a series of prompt designs.

\begin{figure}
    \centering
    \includegraphics[width=\linewidth]{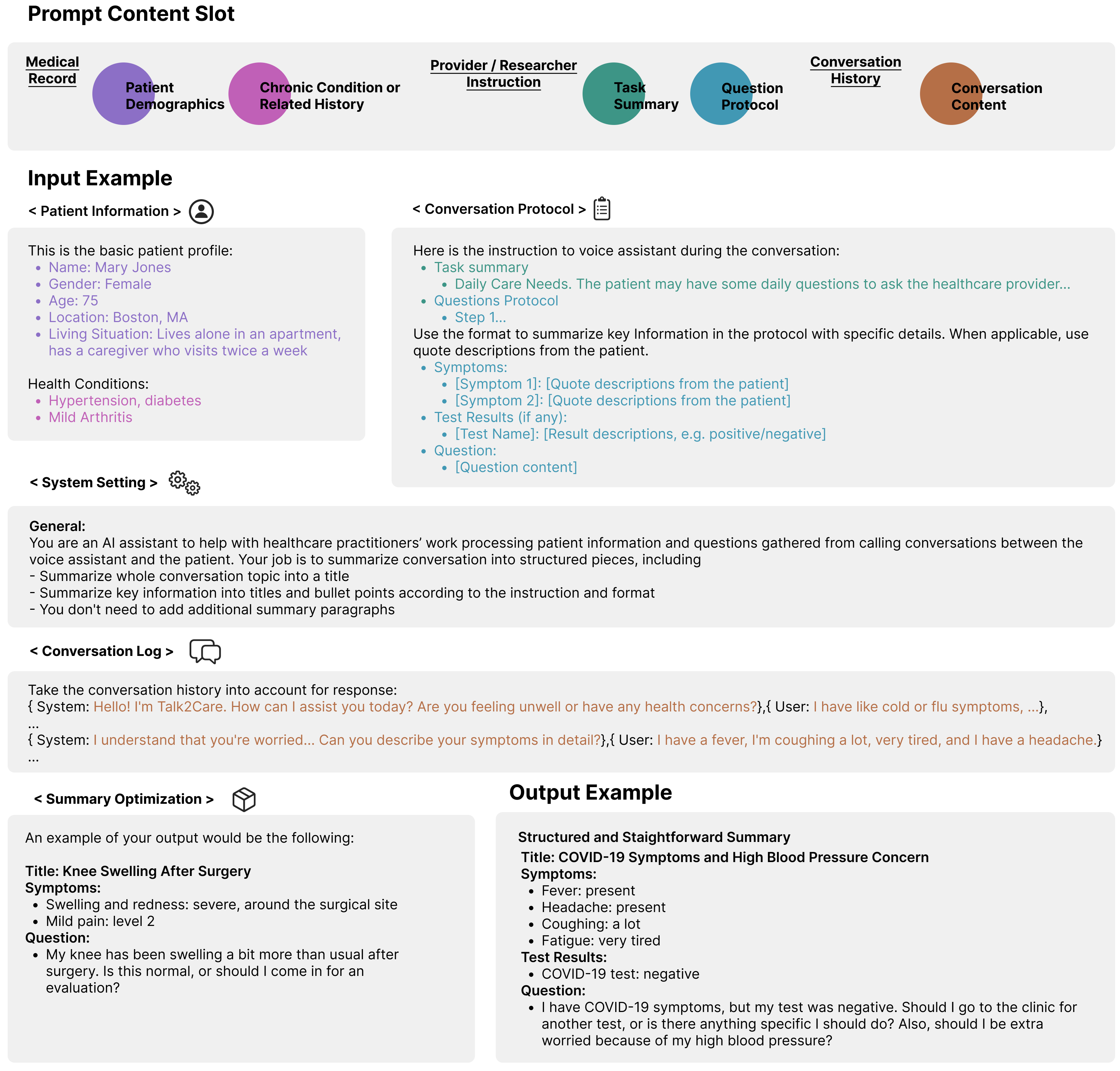}
    \caption{Prompt Design of Patient-VA Conversation Summary for Healthcare Providers. Similar to Figure~\ref{fig:prompt-design-older-adults}, the input prompt consists of five parts: 1) patient information, 2) conversation protocol, 3) system setting, 4) conversation log, and 5) summary optimization. For multi-turn conversation, 5) will be repeated for each round of conversation.
    This figure continues the example of the protocol of daily care.
    }
    \label{fig:prompt-design-healthcare-providers}
\end{figure}

By leveraging provider training tutorials and literature~\cite{AAPD, PIH, Prompt_2023}, our prompt iteration process is close to the one introduced in Section~\ref{subsub:system:patients:question}.
The final prompt design structure, as shown in Figure~\ref{fig:prompt-design-healthcare-providers}, also resembles the question generation for older adults (Figure~\ref{fig:prompt-design-older-adults}), including five factors.
\textbf{(1) Patient Information.}
This is the same information of the older adults from the information database.
\textbf{(2) Conversation Protocol.}
Other than providing the task summary and question protocol, the prompt further emphasizes the key information to help the LLM narrow down its focus and improve the quality of the summary. \imwutreview{Beyond the key information, the protocol also instructs the LLM to highlight any extra communication needs with healthcare providers if requested by users. For example, if the patient shared a particular concern when asked ``is there anything else that I could help?'', the system would present the response as an additional note.} This piece usually comes together with each question protocol. 
\textbf{(3) System Setting.}
This part instructs the role of the AI assistant (as a text summary tool), its responsibilities (i.e., extracting health information and patient questions), and output format (i.e., concise clinical note structure, Section~\ref{subsub:understand:providers:results_LLM}). The specific wordings are shown in Figure~\ref{fig:prompt-design-healthcare-providers}.
\textbf{(4) Conversation Log.}
The complete log of the conversation session between the older adult and the VA.
\textbf{(5) Summary Optimization.}
Through our pilot tests, we noticed the need to adopt an in-context learning paradigm to help the LLM generate high-quality summaries. Therefore, this part shows a summary example to ensure the output is well-structured.

\subsubsection{Information Highlight Module}
\label{subsub:system:providers:highlight}
The healthcare providers in our interviews also mentioned the importance of having the raw conversation log (Section~\ref{subsub:understand:providers:results_LLM}), enabling them to check the raw data when necessary. We design a highlighting feature to save their effort.
The prompt setup is similar to the conversation summary module introduced in Section~\ref{subsub:system:providers:summary}, with the main difference being the system setting (the responsibility is to find and return the important quotes of the older adult) and output optimization (no longer needed).

\subsubsection{Risk Prediction Module}
\label{subsub:system:providers:prediction}
There has been a wide range of research on risk prediction in healthcare (e.g., \cite{kennedy2014delirium,o2015risk,gatt2022review}). We envision our dashboard should also include such a module to help providers allocate limited resources appropriately.
For the completeness of our \projectname{} system, we design an LLM-powered risk prediction module.
The prompt design is also similar to the conversation summary module in Section~\ref{subsub:system:providers:summary}, with the main difference being the system setting (the responsibility is to predict the risk level of the older patient: low, moderate, and high)  and output optimization (ask for the prediction reasoning).
However, we recognize the ethical risks of mis-predictions. Evaluating the prediction performance of the LLM goes beyond the scope of our paper. Additionally, health providers need to implement safety protocols to standardize risk management. 
Therefore, our evaluation of \projectname{} in Section~\ref{sec:evaluation} mainly focuses on the usability of the system design.

\subsubsection{Dashboard Interface}
\label{subsub:system:providers:interface}
We develop our dashboard interface with the React framework.
Our design simplifies the interface of existing EHR systems~\cite{AdvancedMD_2023,Epic} and mainly focuses on the new features we propose.
Figure~\ref{fig:dashboard-interface} presents the continuous example of the older adult's information, following Figure~\ref{fig:prompt-design-older-adults} and Figure~\ref{fig:prompt-design-healthcare-providers}.
Providers can easily navigate to read the conversation summary, the raw conversation logs and highlights, and the communication session history.
For each session, they can take follow-up actions or write notes under the summary.

\begin{figure}
    \centering
    \includegraphics[width=\linewidth]{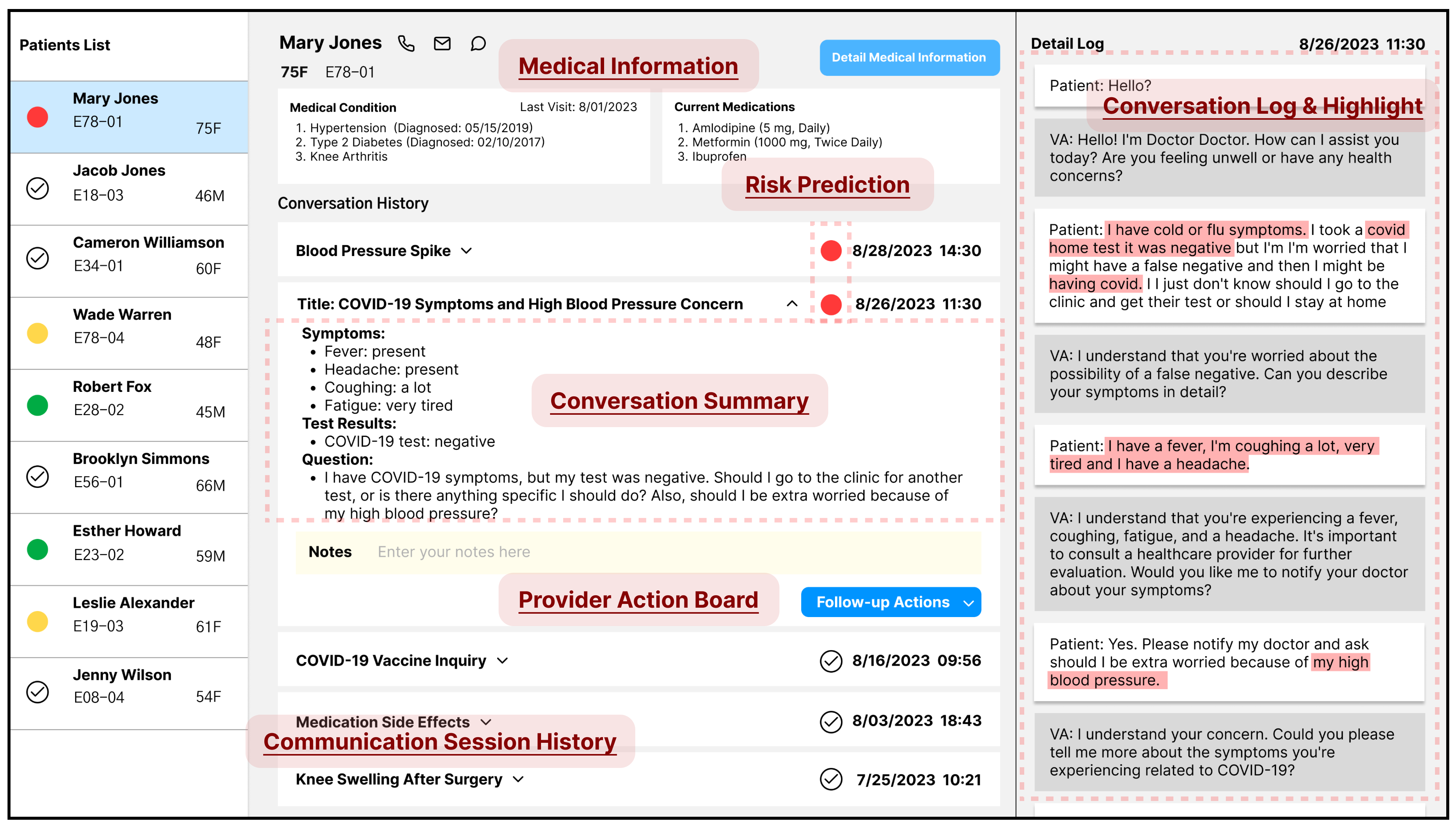}
    \caption{LLM-powered Dashboard Interface Layout for Healthcare Providers. The dashboard presents the conversation summary (center), raw log, and its highlights (right). The colored dot represents the risk prediction (left and center).
    Providers can take specific follow-up actions or write down notes under each conversation summary, after which they can mark one session as done.
    }
    \label{fig:dashboard-interface}
\end{figure}

\subsubsection{Information Database}
\label{subsub:system:providers:database}
The information database of the dashboard is shared with the database of the VA system. It provides the older patient's information, conversation log, and health protocol for prompt input. It also saves providers' action logs or notes for future visualization.

\subsubsection{Interaction Flow}
\label{subsub:system:providers:flow}
Depending on the specific scenarios, the providers first input instructions and protocols into the system, which are used for both the older adult's and the provider's components.
When new conversation sessions are stored in the database, the providers are notified. Using the LLM-powered dashboard, they can check the conversation content, take notes, and make decisions on the appropriate follow-up actions.

Combining the two components together, we aim to build \projectname{} as a complete system to facilitate the communication between home-based older adults and healthcare providers.

\section{Evaluation}
\label{sec:evaluation}
To evaluate the design of our \projectname{} pilot system for stakeholders, we proposed two specific healthcare scenarios to put \projectname{} in realistic settings (Section~\ref{sub:evaluation:scenario}). Then, we conducted two user studies \review{using the two scenarios} with both older adults (Section~\ref{sub:evaluation:older_adults}) and healthcare providers (Section~\ref{sub:evaluation:providers}) to measure the usability of \projectname{} design.


\subsection{Scenario Setup}
\label{sub:evaluation:scenario}
Based on our interview feedback in Section~\ref{sec:understand}, we designed two scenarios corresponding to the communication needs of older adults at home and healthcare providers, one with older adults as initiators (Section~\ref{subsub:evaluation:scenario:daily}), and the other with providers as initiators (Section~\ref{subsub:evaluation:scenario:post-surgery}).

\subsubsection{Scenario A: Daily Care Needs}
\label{subsub:evaluation:scenario:daily}
One of the common communication needs of older adults at home is when they have unexpected health concerns and want to communicate with providers.
We designed a character, Mary, a 75-year-old female patient living alone.
Mary has been coughing and feeling tired since this morning. She suspects it may be related to COVID-19, but the home-test result turns out to be negative. Mary does not want to travel to the clinic unless necessary, because it takes too much time and energy, and she wants to avoid spreading the potential virus. So she wants to ask for advice on going to urgent care or staying at home.
Meanwhile, Mary has high blood pressure, so she is worried about the complex situation and wants to check if there may be any problems.

On the provider side, we designed another character, Tom, a medical office assistant working at a clinic, and his primary responsibility is to answer patient phone calls every day. 
He gets a number of calls from patients asking about various things, such as lab results, prescriptions, random health-related questions, and requests for appointments.
After Tom has all the information from the phone calls, he needs to prioritize them and take the next steps.

\subsubsection{Scenario B: Post-Surgery Follow-up}
\label{subsub:evaluation:scenario:post-surgery}
\review{For providers, one common communication need is to follow up with older patients about their symptoms after hospital discharges.}
Following the character above, John, a 72-year-old male, just went through a small joint surgery on the knee two days ago. He feels good overall, but there is still some pain in the knee. Meanwhile, John has two different kinds of painkillers: aspirin and ibuprofen. He is not sure which one to take.

On the provider side, Emily is a postoperative care nurse working at a busy hospital. Part of her primary responsibility is to follow up with patients after their surgeries, ensuring a smooth recovery process.
Emily has to call a number of patients to ask about their current situation and questions. John is one of Emily's patients, so Emily needs to follow up with him about the knee surgery.
After Emily has all the information, she will prioritize them and take the next steps.

\subsection{User Study \review{1: Conversational Interface} with Older Adults}
\label{sub:evaluation:older_adults}
This user study aims to evaluate the older adult's component of \projectname{}.
Our evaluation results indicate that the pilot system can provide good usability (Section~\ref{subsub:evaluation:older_adults:usability}), help older adults provide more health information (Section~\ref{subsub:evaluation:older_adults:more_info}), and offer better mental support (Section~\ref{subsub:evaluation:older_adults:mental_support}).

\subsubsection{\imwutreview{Methods: Participants and Procedure}}
\label{subsub:evaluation:older_adults:participants}
\imwutreview{With the IRB approval, we recruited the same set of 10 participants from our interview (Table~\ref{tab:study1_demo}). We confirmed that no participants have any hearing or speaking disabilities that may impact the interaction with the VA.}

\begin{figure}
    \centering
    \includegraphics[width=0.5\linewidth]{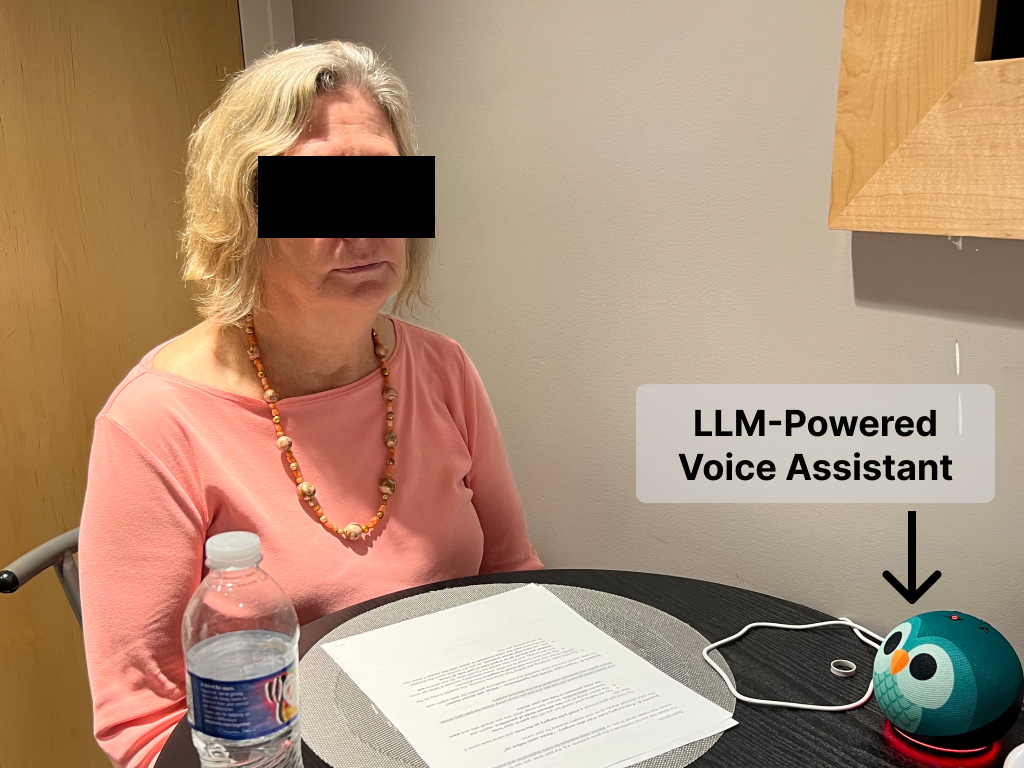}
    \caption{User Study 3 Scenario with an Older Adult Participant.}
    \label{fig:VA-study-example}
\end{figure}

\label{subsub:evaluation:older_adults:design}
We compared \projectname{} against the most common asynchronous communication method by leaving a message to providers.
Specifically, we introduced the two scenarios one by one. After introducing each scenario, older adults were asked to play the role of the character (Mary or John) by writing down (or speaking aloud) the message to leave for providers. Then, they had a conversation with the \review{LLM-powered} VA system \review{we implemented}, with the same goal of conveying the information to providers. \review{Researchers first gave a brief tutorial on the LLM-powered VA and invited participants to test simple interactions such as invocation. Then the older adults were presented with a sheet to remind them of key invocation phrases and scenario setup. During the interactions with the VA, they were encouraged to engage in natural conversations according to their own language style and add details.} After going through the two scenarios, they completed a short evaluation questionnaire and a semi-structured exit interview. Note that the order of the two scenarios was counterbalanced \review{ (\eg participants with odd numbers start with Scenario A and even numbers with Scenario B)}.
Two examples of the conversation log between the older adult and the VA can be found in Appendix \ref{sub:post-surgery-log} and \ref{sub:daily-care-log}.
Figure~\ref{fig:VA-study-example} shows an example of the user study. The study lasted 25-30 minutes. Participants were compensated \$25 for their time.

Our questionnaire includes the System Usability Scale (SUS)~\cite{Brooke_1995_SUS}, the comparison against the basic method of leaving a message, the health support experience, and a few VA-specific experience questions, using a 5-point Likert scale. \review{The SUS is an industrial standard scale to evaluate system usability for users to rate from ``strongly disagree'' to ``strongly agree'' on 10 arguments such as whether the system is easy to understand, their need for technical assistance, and learning difficulty~\cite{Brooke_1995_SUS}.}
In the interview session, we collected user feedback on what they liked and disliked regarding their interaction experience, as well as their future expectations about the system. The interviews were recorded, transcribed, and processed with Thematic analysis by two researchers.
\imwutreview{We summarize our findings as follows.}

\begin{figure}
    \centering
    \includegraphics[width=1\linewidth]{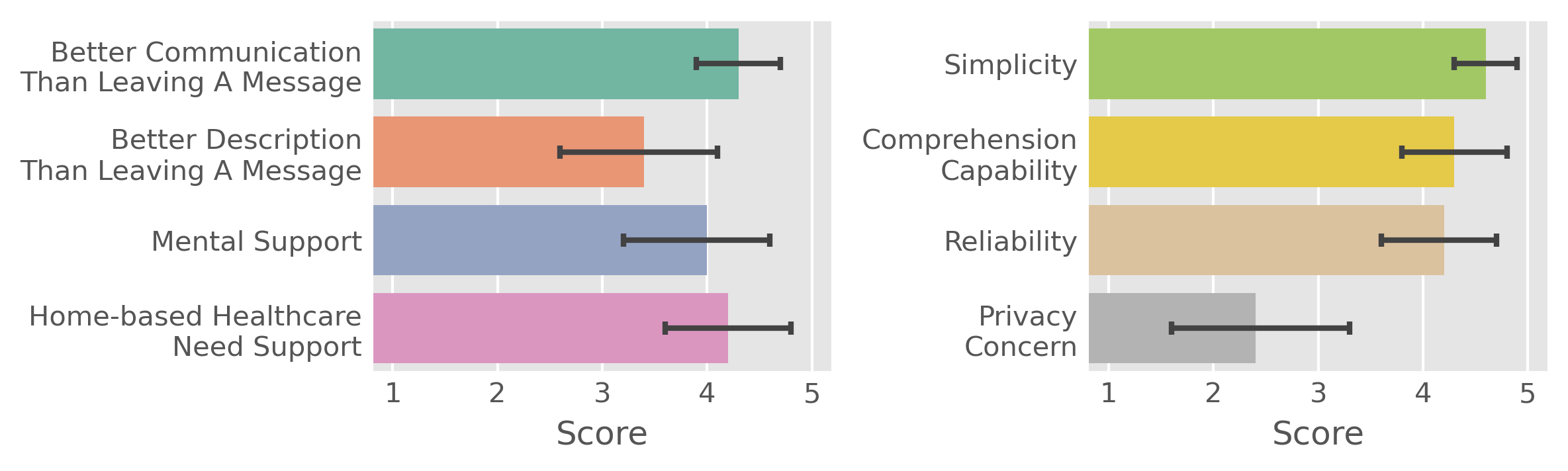}
    \caption{Subjective Ratings on Different Perspectives of The Usability of The VA System. Error bars show the standard deviation. The average SUS score is 75.5$\pm$17.1, indicating good usability.}
    \label{fig:VA-study_questionnaire}
\end{figure}

\subsubsection{\imwutreview{Findings: Good Usability, Description Enhancement, and Mental Support}}
\label{subsub:evaluation:older_adults:usability}
During the study, all participants were able to successfully complete both scenarios following the instructions. They agreed that the two designed scenarios were close to their daily care needs situations (4.2$\pm$1.0 out of 5).
Meanwhile, participants found the VA system easy to use. \imwutreview{The average SUS score was 75.5$\pm$17.1 (out of 100), indicating \textbf{good usability.}}
\pquote{10}{[It's] almost like talking to anybody. I mean, talk to people on the telephone. It's not that different.}
They also found that the VA system's generated questions were simple to understand (4.6$\pm$0.5), and that the system could understand their words accurately (4.3$\pm$0.8). Participants agreed that the system was reliable (4.2$\pm$1.0) and the overall privacy concern was not high (2.4$\pm$1.4).
\imwutreview{Through our analysis of the conversation logs, we find that the LLM can successfully identify varying ambiguous statements from older adults. For example, in Scenario B, participants described the pain as \epquote{5}{my knees are killing me} or \epquote{6} {I have body aches}, and the LLM asked proper follow-up questions about pain level (PP5) or position (PP6).}
\label{subsub:evaluation:older_adults:more_info}
During the study, participants tried the traditional method of leaving a message, so that they could compare it against the method of talking with the VA.
\imwutreview{Our results showed that older adults thought the new method could \textbf{better support their communication} with providers (4.3$\pm$0.7), and \textbf{enhance the descriptions of their health information} (3.4$\pm$1.3).
\pquote{9}{It reminded me of additional symptoms that I went through}.
\pquote{1}{I wouldn't necessarily have thought to include that in my message. [But that is] significant to the doctor.}
}
This was a strong advantage of \projectname{}: With the help of natural and interactive conversation, older adults were able to better communicate and give more health information to providers. 


\label{subsub:evaluation:older_adults:mental_support}
As mentioned in Section~\ref{subsub:understand:older_adults:results_LLM}, older adults thought our system might \imwutreview{\textbf{provide better mental support}}, even though it was intentionally designed to not give specific health advice.
This comment was validated after they used our system. Participants gave high ratings on mental support (4.0$\pm$1.3) and on the support for their home-based healthcare needs (4.2$\pm$1.0).
In particular, P2, previously working as a nurse in a local clinic, provided some insights. ``\textit{A lot of people are sometimes afraid to ask a doctor. They think it's foolish, so I think that gives them the opportunity to ask without being too threatening by picking up a phone and having to talk to a secretary and go through all of that.}''
\imwutreview{This provided an interesting effect of AI-mediated communication, which we will further discuss in Section~\ref{sec:discussion}. }
\subsubsection{\imwutreview{Findings: Future Expectations and Concerns}}
\label{subsub:evaluation:older_adults:expectation}
\imwutreview{Participants had more specific expectations after using the system.}
A dominating direction was to have the system provide more administrative information, integrate with the healthcare system, and help manage their personal health information.
For example, P7 and P9 suggested the VA could provide an estimated time that it takes for the healthcare provider to respond.
Participants also mentioned that the system could play the role of ``\textit{an entry point of the different specialists}'' within the EHR system to minimize the communication effort.

\label{subsub:evaluation:older_adults:concerns}
\review{
During the study, some participants did experience voice assistant failure such as being interrupted and voice recognition failure. \pquote{6}{I think it's a little awkward, the need to have her complete all her sentences. } However, most participants felt confident about adjusting to the interaction mode as long as they have time to practice.  \pquote{6}{Once you get into the rhythm, I think it's okay. } Yet still, some participants thought the reluctance to adaption for older adults might be a barrier, "because it's a big change".  Therefore, a smooth onboarding experience and great usability would be crucial to such an LLM-powered interface for older adults.}
\review{
As mentioned previously, the overall privacy concern was not high (2.4$\pm$1.4) among participants. Interestingly, some participants felt that their privacy were better protected interacting with the VA. \pquote{9}{... as opposed to talking to a nurse that could be a next-door neighbor ... there's a little bit more of a barrier between that and humans seeing the information.}}

\subsection{User Study \review{2}: System Interface Study with Healthcare Professionals}
\label{sub:evaluation:providers}

Our Study 3 evaluated the first half of the \projectname{} system for older adults. This study focused on evaluating our dashboard design's usability with providers. \imwutreview{Our results demonstrate that the pilot system provides good usability (Section~\ref{subsub:evaluation:providers:usability}), and that it can effectively support providers' communication and information processing (Section~\ref{subsub:evaluation:providers:information_analysis}).}
\subsubsection{\imwutreview{Methods: Participants and Procedure}}
\label{subsub:evaluation:providers:participants}
Our study was also IRB-approved. We recruited the same set of 9 healthcare providers from our User Study 2 (Table~\ref{tab:study2_demo}).
All providers have the experience of using one or multiple EHR systems.

\begin{figure}
    \centering
    \includegraphics[width=\linewidth]{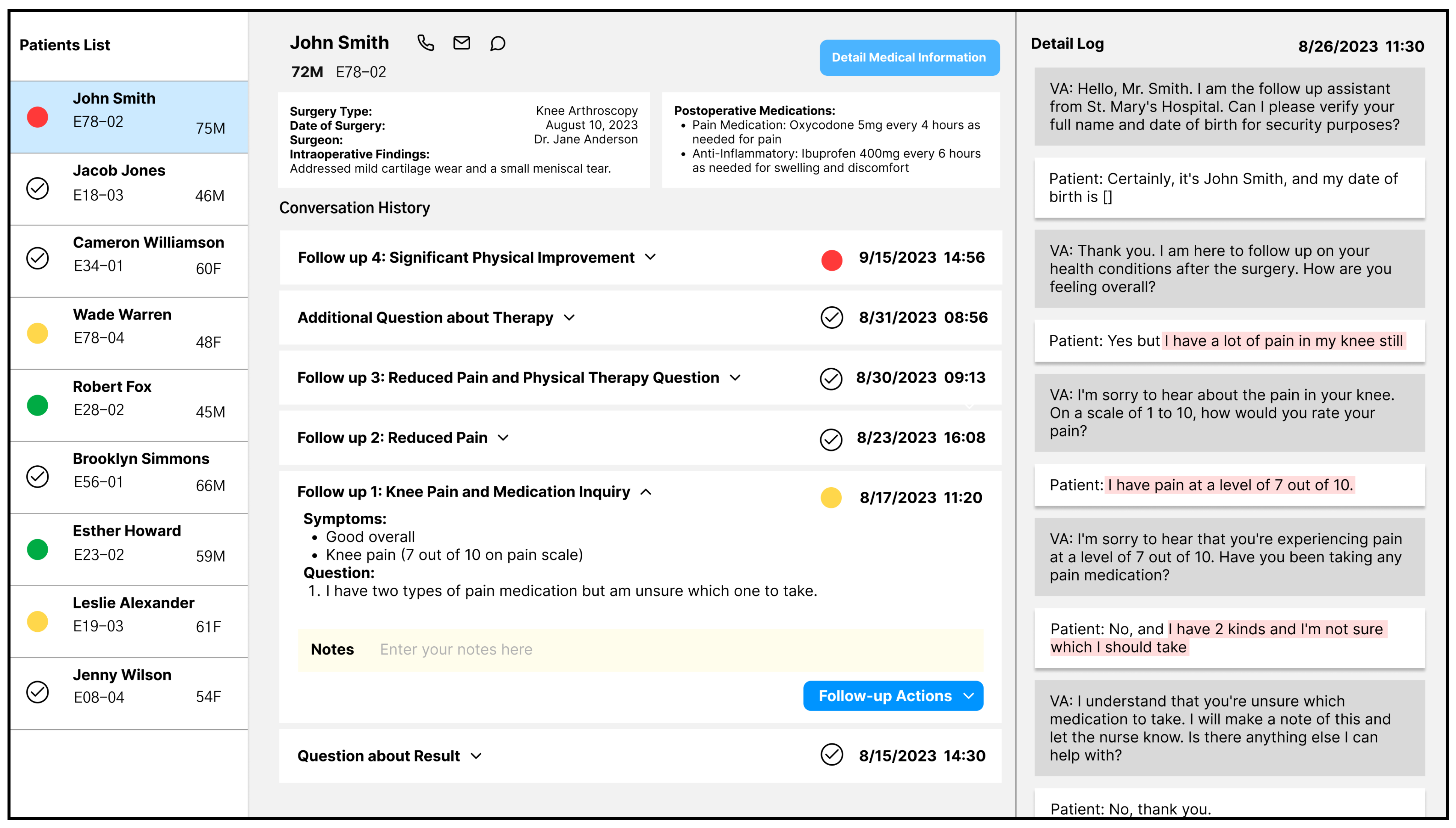}
    \caption{Dashboard Interface Example for Scenario B: Post-Surgery Follow-up. The example for Scenario A can be found in Figure~\ref{fig:dashboard-interface}, with the same interface and different scenario-specific content.}
    \label{fig:post-surgery-dashboard-example}
\end{figure}

\label{subsub:evaluation:providers:design}
Similar to Study 3, we compared \projectname{} against the most common asynchronous communication method of receiving a message from older patients to providers.
We briefly introduced the dashboard interface to providers. After they got familiarized with the system, we introduced the two scenarios to them one by one. Since providers had different responsibilities and specialties, they were not asked to play the role but to suggest the characters (Emily and Tom) about the appropriate actions.
In each scenario, they were first asked to give suggestions on the next steps with the traditional communication method, \ie after receiving patient messages corresponding to the scenario. 
Then, \review{they were invited to share their screen and navigate through different pages in the dashboard interface by clicking, scrolling, or reviewing information to complete the same goal of processing patient information. In the meantime, they were encouraged to share their thoughts about the interface design.}
Note that the dashboard \review{design}of the two scenarios was the same, except for the different content (scenario A: Figure~\ref{fig:dashboard-interface}, scenario B: Figure~\ref{fig:post-surgery-dashboard-example})\review{, where we made up patient information for each scenario where the critical cases were as stated in \ref{sub:evaluation:scenario}}
After going through the two scenarios (with a counterbalanced order across participants), they completed a short evaluation questionnaire and a semi-structured exit interview. 
The study lasted 25-30 minutes. Providers were compensated \$25 for their time.

Our questionnaire includes the NASA Task Load Index (TLX)~\cite{hart1988development}, the System Usability Scale (SUS)~\cite{Brooke_1995_SUS}, the comparison against the basic method of processing messages, the work support experience, and a few dashboard user experience questions. \review{The TLX questions had a 7-point Likert scale to evaluate aspects of the task load. In questions about demand, effort, or frustration, a lower score indicated a lighter workload and was desirable for the system; in those about performance, a higher score indicated a better outcome. All other questions had a 5-point scale, where higher scores indicated better support. } 
In the interview session, we collected providers' feedback on their experience regarding using the dashboard to facilitate their communication with older adults. We also asked about their future expectations and concerns about the system. Like other studies, two researchers processed the interview transcriptions with Thematic analysis.
We summarize our findings below.

\subsubsection{\imwutreview{Findings: Good Usability, More Information, and AI-Mediated Communication}}
\label{subsub:evaluation:providers:usability}
\begin{figure}
    \centering
    \includegraphics[width=1\linewidth]{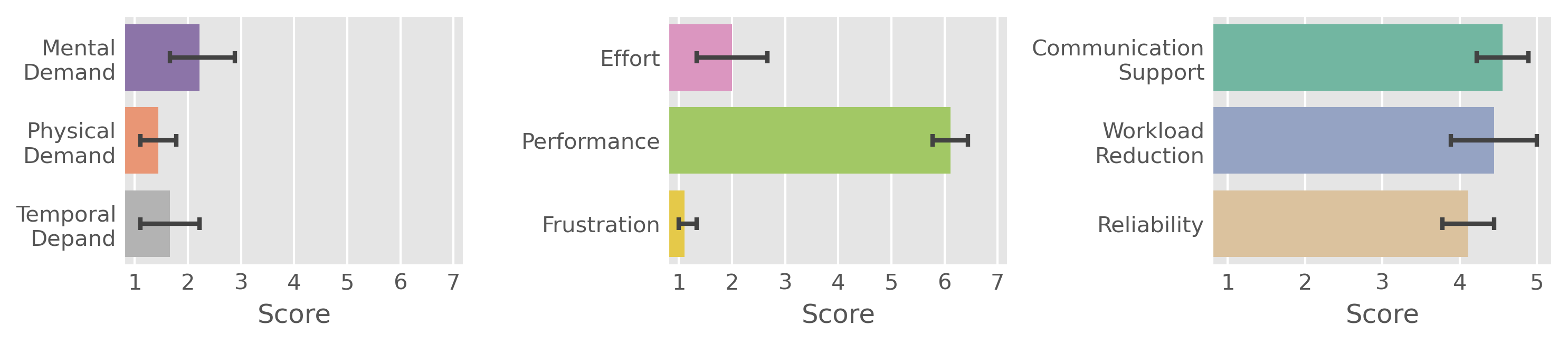}
    \caption{Subjective Ratings on Different Perspectives of The Usability of The Dashboard System Design. The average SUS score is 85.8$\pm$9.8, indicating very good usability.}
    \label{fig:dashboard-study_questionnaire}
\end{figure}

\imwutreview{After navigating through the interface for the two scenarios, providers commented that the system was \textbf{easy to understand and convenient to use}.}
They also agreed that the scenarios and questions were realistic to their actual scenarios, and that the system established a good workflow. 
The system has an overall SUS score of 85.8$\pm$9.8, indicating very good usability.
The NASA TLX results were also consistent with their comments (see Figure~\ref{fig:dashboard-study_questionnaire}), with low ratings on mental demand (2.2$\pm$1.0 out of 7), physical demand (1.4$\pm$0.5), temporal demand (1.7$\pm$0.9), effort (2.0$\pm$1.1), and frustration (1.1$\pm$0.3), as well as high ratings on performance (6.1$\pm$0.6).

\label{subsub:evaluation:providers:communication_support}
Providers also thought the system was helpful in their communication with older adults (4.6$\pm$0.5 out of 5) and offloading their work (4.4$\pm$0.9).
Six participants mentioned cases where the system could save time for them. For example, PP2, PP3 and PP9 were convinced that the VA system could save their time spent on the phone calling. PP1 also mentioned that the whole workflow might also reduce unnecessary outpatient visits.

\label{subsub:evaluation:providers:information_analysis}
\imwutreview{Moreover, also consistent with Study 2, more than half of the providers (5 out of 9) mentioned that the LLM-powered dashboard could \textbf{assist in triage and help prioritize their tasks}, improving efficiency in the process. }
Meanwhile, the summary and highlights could help them organize notes for documentation and speed up the review process (PP2, PP3, PP4, PP7, PP9). This was also backed up by the high ratings on the system reliability (4.1$\pm$0.6).
\imwutreview{Providers also favored leveraging the conversation history for their work in long-term, which could be used for clarification, confirmation, and reminder before a patient's visit (PP1, PP3, PP7).}

\label{subsub:evaluation:providers:more_information}
\imwutreview{In Study 3, patients commented that talking to the intelligent VA system could help them \textbf{reveal more health information }(Section~\ref{subsub:evaluation:older_adults:more_info}). This was also confirmed by providers in Study 4.}
Providers agreed that the summary and log contained more information compared to the commonly seen short messages in the traditional patent portal.
\epquote{9}{When they're writing, they're probably going to put less detail than when they're talking.}
\imwutreview{Compared with some questionnaire practices, they also thought that the information could help narrow down the problem to identify the key issue, and get providers more prepared for future visits.}
With the sufficient information provided, providers also agreed that the communication procedures would be optimized to reduce the back and forth (PP4, PP8).




\label{subsub:evaluation:providers:mental_support}
\imwutreview{Interestingly, providers mentioned that the LLM-powered CA could reduce their mental load of talking with patients by playing \textbf{the intermediate communication role}.}
\imwutreview{\epquote{5}{[Talking with patients is] pretty soul-sucking, ... and [the nurse] has to be good at keeping their temper and managing people's expectations...}}

\subsubsection{\imwutreview{Findings: Future Expectations and Concerns}}
\label{subsub:evaluation:providers:expectation}
In addition to the positive comments, providers also brought up more expectations.
Four providers mentioned the need to integrate the system with EHR systems, which was also brought up by older adults in Study 3.
Also consistent with older adults' expectations, providers mentioned that the system could offer pre-set valid suggestions and explanations from providers.
\imwutreview{For instance, examples for rating pain levels should come together with the question querying pain level as pre-set explanations (PP7). Meanwhile, we only covered two protocols for the two scenarios, and providers expected a future intelligent system could incorporate more detailed protocols customized to medical history and key symptoms, and thus collect more detailed health information (PP2, PP3, PP7). For example, PP3 suggested having different protocols for different surgery types and following up with bleeding-related questions, such as recent bleeding time and any effective interventions.}
With the concerns about AI making mistakes, some providers also suggested that the system should enable them to edit content to correct errors, if any. \imwutreview{When \epquote{8}{the human component comes in}, they feel they would not\textit{ ``miss anything''.}}

\subsection{\imwutreview{Finding Summary}}
\label{sub:userstudy:summary}
\imwutreview{
In summary, we present the following key findings from our user studies, guiding our discussion in Section \ref{sec:discussion}.
\begin{enumerate}
    \item \projectname{} has great accessibility and usability for older adults and their healthcare providers.
    \item The LLM-powered CA could enrich the information communicated and provide mental support for older adults.
    \item The LLM-powered dashboard could promote provider efficiency by offloading and supporting prioritization.
    \item Users identify ethical and privacy concerns as well as desirable features in such LLM-powered systems.
\end{enumerate}
}

\section{Discussion}
\label{sec:discussion}
User Studies 3 and 4 demonstrate the effectiveness of the LLM-powered \projectname{} to facilitate communication between home-based older adults and healthcare providers. In this section, we discuss the effects of AI-mediated communication (Section~\ref{sub:discussion:communication}), the potential of deployment \projectname{} in real-world scenarios (Section~\ref{sub:discussion:deployment}), and the important ethical concerns and safety considerations of \projectname{} (Section~\ref{sub:discussion:ethics}).
We also discuss the limitations of our current work (Section~\ref{sub:discussion:limitation}).

\subsection{AI-Empowered Healthcare Work}
\imwutreview{Our interviews and user studies revealed that the LLM-powered system has the strong potential to fill in the gap between older adults' expectations of their healthcare providers and their reality. Firstly, with pre-set protocols and explanations, LLM could satisfy some of the communication needs of patients if their providers are not immediately available, improving responsiveness and supporting triage. Furthermore, LLMs could also address the limitations of existing communication technologies. With given protocols and interactive conversations, they provide higher adaptability and efficiency for the asynchronous communication. Thus, LLM-powered CAs could promote the level of details of asynchronous communication.}
For healthcare providers, such AI could also lighten their workload in patient-provider communication by taking over basic or repetitive tasks, such as answering phone calls according to a protocol and finding key information in a conversation. Thus, AI might be able to help patients and providers prioritize their communication tasks and maximize communication outcomes within limited availability.
Future designers and system builders could utilize AI to work on the tasks in healthcare communication that are time-consuming but could be well structured with instructions.

\label{sub:discussion:communication}
In multiple stages of our user studies, older adults and providers mentioned how the role of the LLM-powered system can support their communication needs and reduce mental load.
Even if they do not receive specific health advice, older adults still feel mentally supported when talking to our VA system (Section~\ref{subsub:evaluation:older_adults:mental_support}), and some might feel less judged compared to when talking to a real healthcare provider (Section~\ref{subsub:understand:providers:results_LLM}).
For providers, other than saving their effort and time, such an intelligent system can also reduce their mental load for patient communication (Section~\ref{subsub:evaluation:providers:mental_support}).
These results suggest the potential of AI as a mediator role in human-human communication, which can help to take over some less favorable parts during the \review{communication} and improve both patients' and providers' experience.
\review{At the same time, as the AI-powered system promotes the professionalism of providers during communication, such an AI mediator also minimizes harm to patient-provider relationships. }
This work could serve as an example to support the promising future of AI-mediated \review{healthcare} communication~\cite{Mieczkowski2021-mq, hohenstein2018ai}.
For more personal communication scenarios, such as sharing health information across a family, it remains \review{an} open question on how such an AI mediator may affect human relationships.


\imwutreview{Despite our limited number of participants, our participant groups have a rather wide range of medical conditions (patients) or areas of expertise (providers), covering different communication needs and challenges.
Meanwhile, our patient participants' medical conditions are quite common among the general patient population (\eg migraine, hypertension).
Therefore, we envision that our findings could generalize to patient-provider communication for older adults and potentially other patient populations. We discuss more in future work in Section~\ref{sub:discussion:limitation}.
}

\subsection{\imwutreview{Necessary Functions before Real-World Deployment}}
\label{sub:discussion:deployment}
Our studies suggest that \projectname{} has a strong potential to be deployed in the real world. However, there are a number of features that \projectname{} needs to support and be evaluated before deployment (Section~\ref{subsub:evaluation:older_adults:expectation} and \ref{subsub:evaluation:providers:expectation}).

\subsubsection{\imwutreview{Integration with EHR Systems.}}
First, both older adults and providers suggested the idea for our system to be integrated with EHR systems, which could enable our system to provide more personalized conversation with older adults with more validated and detailed question protocols, update their health information in the system more seamlessly, support providers with a smoother working flow, and further facilitate the communication between older adults and multiple providers.
\imwutreview{Admittedly, integration with EHR may come across several obstacles, including demand for budget and resources, handling sensitive information, and interoperability issues~\cite{heart_review_2017, cifuentes_electronic_2015,weng_using_2012}.
We envision such an integration could potentially start at a specialized clinic and include minimal sensitive data.}

\subsubsection{\imwutreview{Incorporating Long-term Conversation History.}}
\imwutreview{Another opportunity to develop a personalized experience is to leverage conversational history in the interaction between patients and LLMs. In our study, since we are using a scenario-based user study setting, we mainly leveraged the conversation history on a per-session scale. Our results have revealed that short-term history (within the same round of conversation) helps the system ask key questions, probe for more health details, and keep the conversation brief. Beyond the scope of the interactions during the study, participants also expressed a strong interest in the possibility of using long-term history (between different conversations) to personalize their experiences. This suggests another direction for our future LLM system.
For example, if the user mentions a symptom of pain one day, the LLM CA could follow up in the next few days and inquire if the pain persists.
In our current implementation, we have enabled a basic long-term memory mechanism in our system by keeping all conversation records.
However, we also recognize the technical difficulties when integrating long-term memory. As pointed out by existing work in LLMs and natural language processing, there are restrictions to the number of tokens input to the LLMs~\cite{noauthor_openai_2023}, limiting the length of conversation history texts that could be included in LLM prompting~\cite{baeKeepMeUpdated2022, leePromptedLLMsChatbot2023, wangAugmentingLanguageModels2023}. Potential solutions may include using conversation summaries instead of original transcripts for long-term use.
Still, future work could further explore the technical solutions to enhance long-term system memory and their effectiveness. 
}



\subsubsection{\imwutreview{Establishing the Boundary of System Capability.}}
\imwutreview{
Given the complexity of medical conditions, such systems should have flexibility in handling various health concerns. Some functions of the current system need to be automated, such as selecting the question protocol based on the varying scenarios. The protocols could also be more specific to patients' health conditions or adopt more complicated logic compared to our scenario-based studies. In addition, failure of the CA such as voice recognition issues (Section \ref{subsub:evaluation:older_adults:concerns}) ~\cite{murtarelli_conversation-based_2021,chaves_how_2021}, should also be addressed to ensure the information accuracy. }

\imwutreview{However, even with close integration with medical and conversation history, such LLM-powered systems may still come across complicated scenarios that challenge their capabilities in real-world deployment.
For example, patients may respond with a mixture of symptoms and questions that do not fit into a pre-set protocol. 
Patients may also ask about a novel or rare term about which the LLM lacks knowledge.
This calls for the LLM's higher flexibility in handling various health concerns, which is an active research topic.
Coming across these capability boundaries, the LLM should respond with recognition of the limitations, such as ``Sorry, I don't know that one, so I will record your questions and have your provider get back to you.''.}

\subsection{Ethical\review{, Privacy} and Safety Concerns of LLM-powered Healthcare Communication}
\label{sub:discussion:ethics}
Although \projectname{} shows a number of promising advantages, ethical\review{, privacy} and safety concerns must be addressed adequately and thoroughly before any real-world deployment.
Our participants mentioned a range of ethical concerns that need to be considered in future work (Section~\ref{subsub:understand:older_adults:results_concerns} and \ref{subsub:understand:providers:results_concerns}).

\subsubsection{\imwutreview{Ethical Concerns: Ensuring Reliability and Human Involvement}}
\projectname{} is intentionally designed to not provide health advice due to the concern about AI reliability in healthcare~\cite{butow2020using, iliashenko2019opportunities, morley2020ethics}. Nevertheless, it can still make mistakes at multiple stages of communication, such as misunderstanding older adults' meaning of a sentence, misleading older adults when providing their health information, or mis-predicting the severity level so that providers' priorities are biased.
\imwutreview{Potential solutions include having a separate sanity-check AI agent (thus forming a group of LLM-based agents), requesting human attention when the system is uncertain, or a combination of both.}

\imwutreview{Similarly, participants also reported reliability concerns regarding LLM's risk predictions, summaries, and highlights.
Thus, we emphasize that the human counter-check in our dashboard interface is a crucial step in the proposed workflow, and providers should be informed of the potential risks of LLM results. Meanwhile, patients could also have access to the transcriptions and LLM outputs for double-check. Furthermore, we recommend future work to systematically evaluate the reliability of such predictions and interpretations, thus guiding real-world deployment.}

\imwutreview{Meanwhile, the impact of such a remote communication system on in-person outpatient visits and patient-provider relationship, as mentioned in participants' concerns, is also unclear.}
Although providers can ask patients to come in for more tests after the system mediates the communication, this system may reduce older adults' desire to leave home and increase health risks. ~\review{Similarly, given earlier work on deception and social CAs~\cite{sharkey_we_2021}, it is possible that this system could lead to the social isolation of some older adults if they communicate more with the system and less with their family members or providers directly.} \imwutreview{Future LLM-powered systems should further explore the essential patient-provider interactions to where direct synchronous patient-provider interactions are crucial.}

\subsubsection{\imwutreview{Privacy Concerns: Informed Consent and Interventions}}
In addition, in future deployment, privacy risk is another important concern of the LLM-powered system, as the system may collect patients' personal information.

As mentioned by participants in interviews (Section \ref{subsub:understand:older_adults:results_concerns} and \ref{subsub:evaluation:older_adults:concerns}), future designers and developers should refer to regulations such as HIPPA and existing secured telehealth systems to enable secured access to the database~\cite{rights_ocr_hipaa_2015, hall_for_2014}, and could potentially utilize natural language processing (NLP) methods to filter critical identifiable information. 
\imwutreview{Meanwhile, in deployment, patients should be fully informed of the data collection of the system.
We suggest a system should at least cover the following two steps.
Firstly, the system should explicitly explain the data usage and privacy concerns, especially in cases when patients have a lower technical literacy. The system can use either traditional written consent forms, explanations by health providers, or explicit reminders from the system.
Secondly, during regular human-CA interactions, the system should regularly remind the patients of risks when sensitive data is collected and suggest practical alternatives.} 

Much future work is needed to carefully address these concerns to ensure a reliable, safe, and robust healthcare communication system. 





\subsection{\imwutreview{Limitations and Future Work}}
\label{sub:discussion:limitation}
There are a few limitations in our work.
First, our sample sizes for the user studies are limited (10 older adults and 9 healthcare providers). \review{Also, since our participants are recruited from the university participation pool, some of them may have higher education levels, digital literacy, and a better understanding of LLMs than other older adult groups. Therefore, the} results obtained from our study may not generalize to other populations. 
\imwutreview{
We also acknowledge the limited number and occupation of our provider participants. We selected the provider profiles to focus on clinical communication needs and patient concerns, including other healthcare providers who have more frequent communication and patient management experience may enrich the study insights. However, the provider participants are mainly physicians instead of nurse practitioners or support staff. Thus, we suggest future work include a wide range of both patient and provider populations to further study the generalizability of such systems in real-world deployment.
}

Second, our current system is a proof-of-concept prototype. The protocols we used in the study need manual effort and can be more detailed. As we discussed in Section~\ref{sub:discussion:deployment}, it needs more future work to achieve a smooth deployment. 
\imwutreview{For example, although our findings suggest that LLMs could enable the system to be customized for each patient, we did not integrate patient information and provider instructions in our user study in a large scale. Resource limitations, privacy, and ethical concerns would arise when scaling up the system.}
\imwutreview{Third, the evaluation of the system mainly focused on system usability and LLM opportunities, but we did not evaluate the performance of different modules in \projectname{}, such as the quality of the generation question or summary, or the accuracy of the risk prediction. Before scaling up the system, there should be a comprehensive and systematic evaluation of the LLM performance in the different modules to ensure reliability and minimize risk to patients.} 





\section{Conclusion}
\label{sec:conclusion}
In this paper, we designed and implemented \projectname{}, a novel LLM-powered pilot system to facilitate communication between older adults at home and healthcare providers.
To better understand the communication challenges and opportunities of LLM-based systems, we first conducted two semi-structured interview studies, one with older adults (N=10), and the other with healthcare providers (N=9).
The interview results reveal the need for an AI-based asynchronous communication system and shed light on our design of \projectname{}.
Our pilot system consists of two components. The patient component is an LLM-powered VA system that generates high-quality questions to collect health information from older adults and send it to providers for further decision-making.
The provider component is a dashboard that summarizes and highlights key information from the patient-VA conversation to save providers' time and assist their analysis process.
We then conducted two more user studies to evaluate \projectname{}.
Our results showed that \projectname{} may address the challenges mentioned by patients and providers, facilitate their communication, enrich the health information shared by older adults, and reduce the efforts of providers.
Our work serves as an example of LLMs' application in the intersection of healthcare and interpersonal communication.


\bibliographystyle{ACM-Reference-Format}
\bibliography{
bib/related-work-1,
bib/related-work-2,
bib/related-work-3,
bib/related-work-intro,
bib/related-work-other,
bib/review,
bib/review-imwut
}

\newpage

\appendix

\section{Appendix: Interview Scripts}

\subsection{Interview Scripts for Older Adults}
\label{sub:appendix-script-older}

We begin with a brief introduction of our research and study procedure. 

\textbf{Background}
\begin{enumerate}
	\item Could you please tell me a bit more about your health conditions, such as any chronic diseases, surgeries or accidents?
	\item Could you please tell me about your home setting, like do you live alone or with family?
\end{enumerate}

\textbf{Recent Healthcare Visit Experience}

\begin{enumerate}
	\item Could you please give a recent example of your healthcare experience with a provider, such as visiting a doctor?
 \begin{enumerate}
     \item What difficulties and inconveniences did you encounter?
 \end{enumerate}

\end{enumerate}

\textbf{Patient-Provider Communication}

\begin{enumerate}
	\item In this process, could you please tell me more about what and when you communicated with a healthcare provider at home or seeked their support? [What questions and needs do you have?]
 \begin{enumerate}
     \item If the health providers are not available, what’s your alternative? Any other solutions you refer to?
 \end{enumerate}

\end{enumerate}

\textbf{Personalization/Conversation}
\begin{enumerate}
    \item In this process, do you have any personalized questions or concerns that you feel can only be addressed by communicating with a healthcare provider(e.g. but not by a friend or online searching)?
\end{enumerate}

\textbf{Attitudes of VA for patient-provider communication}

\begin{enumerate}
    \item Do you know about voice assistants such as Alexa? What about AI? How about ChatGPT? If not, I can make a brief introduction. 
    \item In your case, if there is such an AI voice assistant to help you communicate with a healthcare provider by …, do you think it will be helpful? [What do you think about it? What do you like/dislike about it? What other features would you like or dislike?]
\end{enumerate}

\textbf{Closing Question}
\begin{enumerate}
    \item Do you have any questions for us about the study? 
\end{enumerate}

\subsection{Interview Scripts for Healthcare Providers}
\label{sub:appendix-script-provider}
We begin with a brief introduction of our research and study procedure. 

\textbf{Background}
\begin{enumerate}
    \item Could you please tell me a bit more about your work responsibilities communicating with patients at home, such as monitoring their health conditions or answering patient questions?
\end{enumerate}

\textbf{Patient-Provider Communication}

\begin{enumerate}
	\item In this process, could you please tell me more about what difficulties and challenges you encountered communicating with patients, especially older adults?
		\item For your difficulties, did you seek any support from technologies?
	\begin{enumerate}
			\item Automated calling systems?
			\item Pre-designed surveys?
	\end{enumerate}

\end{enumerate}

\textbf{Attitudes of AI for patient-provider communication}

\begin{enumerate}
    \item Do you know about voice assistants such as Alexa? What about AI? How about ChatGPT? If not, I can make a brief introduction. 
    \item In your case, if there is such an AI voice assistant to help you gather information from your patients by [referring to their difficult cases mentioned above]\\
        Or: In your case, if there is such an AI that summarizes the conversation and highlights the key information, \\
        Do you think it will be helpful?
\begin{enumerate}
	\item What do you think about it?
	\item What do you like/dislike about it?
	\item What other features would you like or dislike?
\end{enumerate}
    \end{enumerate}

\textbf{Closing Question}
\begin{enumerate}
    \item Do you have any questions for us about the study?
\end{enumerate}

\section{Appendix: Conversation History}

\subsection{Post-Surgery Example of Conversation Log Between The Older Adult and Our Voice Assistant}
\renewcommand{\arraystretch}{1.3}
\label{sub:post-surgery-log}

\begin{table}[h]
    \centering
    \begin{tabular}{p{0.2\linewidth} | p{0.75\linewidth}}
\hline \hline
        \textbf{Voice Assistant} & Hello! I'm Talk2Care. How are you feeling after your surgery? Any discomfort or concerns? \\ \hline
        \textbf{Patient} & I'm feeling good overall, thanks for asking. But I have a little pain. \\ \hline
        \textbf{Voice Assistant} & I'm glad to hear that you're feeling good overall. I'm sorry to hear about the pain. On a scale of 1 to 10, how would you rate your pain? \\ \hline
        \textbf{Patient} & I'm not sure. Can you give some examples of how to rate my pain? \\ \hline
        \textbf{Voice Assistant} & On a scale of 1 to 10, with 1 being no pain and 10 being the worst pain imaginable, how would you rate your current pain level? \\ \hline
        \textbf{Patient} & I would probably rate 2. I think it's not too severe. \\ \hline
        \textbf{Voice Assistant} & I understand. On a scale of 1 to 10, with 1 being mild and 10 being severe, you rate your discomfort as a 2. Is that correct? \\ \hline
        \textbf{Patient} & Yes, that's correct. \\ \hline
        \textbf{Voice Assistant} & Thank you for letting me know. Is the pain more towards the mild side or is it more severe? \\ \hline
        \textbf{Patient} & I would say it's more mild, and actually I'm not sure about my painkiller. I have 2 different kinds, but I kind of want to ask my doctor which one I should use. \\ \hline
        \textbf{Voice Assistant} & I understand that you have two different painkillers. Would you like me to pass this information to the doctor and have them advise you on which one to use? \\ \hline
        \textbf{Patient} & Yes, please. And also I think one of them is aspirin and the other one is like a paste that I can put on my skin. Maybe you can ask my doctor about that? \\ \hline
        \textbf{Voice Assistant} & Thank you for letting me know about the medications. I'll make sure to pass that information along to your doctor. Is there anything else I can help you with? \\ \hline
        \textbf{Patient} & No, thank you. \\ \hline
        \textbf{Voice Assistant} & Goodbye! \\ \hline \hline
    \end{tabular}
\end{table}
\newpage

\subsection{Daily Care Example of Conversation Log Between The Older Adult and Our Voice Assistant}
\label{sub:daily-care-log}
\begin{table}[h]
    \centering
    \begin{tabular}{p{0.2\linewidth} | p{0.75\linewidth}}
        \hline \hline
        \textbf{Patient} & Hello? \\ \hline
        \textbf{Voice Assistant} & Hello! I'm Talk2Care. How can I assist you today? Are you feeling unwell or have any health concerns? \\ \hline
        \textbf{Patient} & I have like cold or flu symptoms, I took a covid home test it was negative, but I, I'm not, I'm worried that I might have a false negative and then I might be having covid. I, I just don't know should I go to the clinic and get their test or should I stay at home. \\ \hline
        \textbf{Voice Assistant} & I understand that you're worried about the possibility of a false negative. Can you describe your symptoms in detail? \\ \hline
        \textbf{Patient} & I have a fever, I'm coughing a lot, very tired, and I have a headache. \\ \hline
        \textbf{Voice Assistant} & I understand that you're experiencing a fever, coughing, fatigue, and a headache. It's important to consult a healthcare provider for further evaluation. Would you like me to notify your doctor about your symptoms? \\ \hline
        \textbf{Patient} & Yes, please notify my doctor and ask should I be extra worried because of my high blood pressure. \\ \hline
        \textbf{Voice Assistant} & I understand your concern. It's important to monitor your symptoms closely. Would you like me to inform your healthcare provider about your symptoms? \\ \hline
        \textbf{Patient} & Inform my doctor, thank you. \\ \hline
        \textbf{Voice Assistant} & I have informed your doctor about your symptoms. They will contact you directly for further guidance. Is there anything else I can assist you with? \\ \hline
        \textbf{Patient} & No, thank you. \\ \hline
        \textbf{Voice Assistant} & Goodbye! \\ \hline \hline
    \end{tabular}
\end{table}
\renewcommand{\arraystretch}{1.0}


\end{document}